\documentclass[10pt,journal,compsoc]{IEEEtran}
\usepackage{graphicx}
\usepackage{tikz}
\usepackage{comment}
\usepackage{enumitem}
\usepackage{amsmath,amssymb} 
\usepackage{ntheorem}
\usepackage{color}
\usepackage{comment}
\usepackage{booktabs}
\usepackage{multirow}
\usepackage{verbatim}
\usepackage{bbm}

\usepackage{listings}
\usepackage{colortbl}
\usepackage{indentfirst}
\usepackage{subfigure}
\usepackage{floatrow}
\usepackage{wrapfig}
\usepackage{lipsum}
\ifCLASSOPTIONcompsoc
  \usepackage[nocompress]{cite}
\else
  \usepackage{cite}
\fi
\ifCLASSINFOpdf
\else
\fi
\begin{document}
%
\title{Provably Uncertainty-Guided Universal Domain Adaptation}
%
%
%
%

\author{
Yifan Wang, Lin Zhang, Ran Song,  Paul~L.~Rosin, Yibin Li, and Wei Zhang

\IEEEcompsocitemizethanks{\IEEEcompsocthanksitem Yifan Wang, Lin Zhang, Ran Song, Yibin Li, and Wei Zhang are with the School of Control Science and Engineering at Shandong University, China.
\IEEEcompsocthanksitem Paul~L.~Rosin is with the School of Computer Science and Informatics, Cardiff University, Cardiff, UK.
\IEEEcompsocthanksitem Corresponding author: Wei Zhang (Email: davidzhang@sdu.edu.cn)}
}

\IEEEtitleabstractindextext{%
\begin{abstract}
Universal domain adaptation (UniDA) aims to transfer the knowledge from a labeled source domain to an unlabeled target domain without any assumptions of the label sets, which requires distinguishing the unknown samples from the known ones in the target domain. A main challenge of UniDA is that the nonidentical label sets cause the misalignment between the two domains. Moreover, the domain discrepancy and the supervised objectives in the source domain easily lead the whole model to be biased towards the common classes and produce overconfident predictions for unknown samples. To address the above challenging problems, we propose a new uncertainty-guided UniDA framework. Firstly, we introduce an empirical estimation of the probability of a target sample belonging to the unknown class which fully exploits the distribution of the target samples in the latent  space. Then, based on the estimation, we propose a novel neighbors searching scheme in a linear subspace with a $\delta$-filter to estimate the uncertainty score of a target sample and discover unknown samples. It fully utilizes the relationship between a target sample and its neighbors in the source domain to avoid the influence of domain misalignment. Secondly, this paper well balances the confidences of predictions for both known and unknown samples through an uncertainty-guided margin loss based on the confidences of discovered unknown samples, which can reduce the gap between the intra-class variances of known classes with respect to the unknown class. Finally, experiments on three public datasets demonstrate that our method significantly outperforms existing state-of-the-art methods. 

\end{abstract}

}

\maketitle

\IEEEdisplaynontitleabstractindextext

%
\IEEEpeerreviewmaketitle

\IEEEraisesectionheading{\section{Introduction}\label{sec:introduction}}

%
%
%
%

\begin{figure*}[t]
  \centering
   \includegraphics[width=1\linewidth]{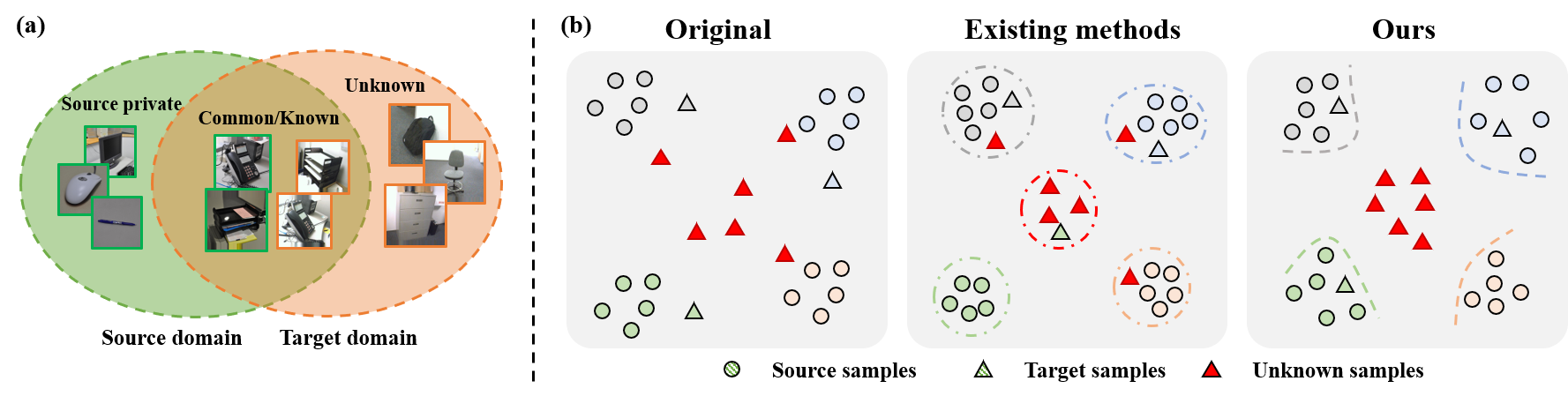}
   \vspace{-1mm}
   \caption{ {(a) Illustration of Universal domain adaptation. (b) Comparison of UniDA methods. The distribution of embedings in the original feature space are highly misaligned because of the domain discrepancy. The existing neighborhood-based methods has bad performance on matching the samples from the two domains. Our method can reduce the influence of domain misalignment and find the unknown samples reliably.}}
   \label{intro}
\end{figure*}

Unsupervised domain adaptation (UDA)~\cite{ganin2015unsupervised,kang2019contrastive,long2016unsupervised,xiao2021dynamic,saito2021ovanet,liu2019separate} aims to transfer the knowledge from a labeled source domain to a fully unlabeled target domain. Early work of UDA, now usually called closed-set DA (CDA)~\cite{gong2013connecting,zou2018unsupervised,na2021fixbi}, assumes that the label sets of the source domain and the target domain are identical. The knowledge transfer between the two domains is thus relatively straightforward due to the identical label sets, but the applications of CDA are limited in real-world scenarios. Subsequently, quite a few methods have been proposed to handle UDA problems with more relaxed assumptions. 
Partial-set DA~(PDA)~\cite{cao2018partial,cao2019learning,zhang2018importance,liang2020balanced} assumes that the label set of the target domain is a subset of that of the source domain.
On the contrary, Open-set DA~(ODA)~\cite{panareda2017open,saito2018open,liu2019separate} assumes that classes in the source domain are all present in the target domain but some classes in the target domain are \emph{unknown} in the source domain.
Open-partial DA~(OPDA)~\cite{saito2021ovanet,li2021DCC,fu2020learning} introduces private classes for both domains respectively, where the private classes in the target domain are defined as unknown classes.
As illustrated in Fig.~\ref{intro}(a), Universal DA~(UniDA)~\cite{bucci2020effectiveness,saito2021ovanet,li2021DCC,wang2024exploiting} is the UDA with the most general setting, where no prior knowledge is required on the label set relationship between domains. A main challenge of UDA is the domain misalignment caused by the biased and less-discriminative embedding. The misalignment may mislead the knowledge transfer and result in an incorrect classification. In the UniDA, the label sets of two domains are not exactly overlapped, which magnifies the domain bias. Thus, it is important to distinguish the unknown target samples to reduce the influence of the domain misalignment.

To address the above problem in UniDA, a popular type of methods~\cite{saito2020universal,li2021DCC,chen2022geometric} are to complete the alignment between samples in common classes of both source and target domains and push the unknown samples away from common classes. For instance, Saito \textit{et al.}~\cite{saito2020universal} proposed a prototype-based method to move each target sample either to a prototype of a source class or to its neighbors in the target domain. Li \textit{et al.}~\cite{li2021DCC} solved this problem by replacing the classifier-based framework with a clustering-based one which exploited the intrinsic structure of samples and thus increased the inter-sample affinity in each cluster. Chen~\textit{et~al.}~\cite{chen2022geometric} proposed a geometric anchor-guided adversarial and contrastive learning framework with uncertainty modeling, which achieved the state of the art (SOTA) by a global joint local domain alignment strategy.

However, without any prior knowledge about unknown samples and source private classes,  approaches of completing the alignment \cite{saito2020universal,li2021DCC,chen2022geometric} between two domains are risky, and can even magnify the misalignment. 
As illustrated in the leftmost part of Fig.~\ref{intro}(b), since samples in the unknown class are not identical, the intra-class affinity of the unknown class is much lower than that of any known class especially when the unknown set is large.
This means that the affinity between two samples in the unknown class can be even lower than that between an unknown sample and a known sample. In addition, due to the less-discriminative embedding, the affinity between a known sample and unknown samples can be greater than that between it and samples in the same source class.
Consequently, as illustrated in the middle part of Fig.~\ref{intro}(b), some unknown samples are easily pushed closer to one of the source classes incorrectly and some known samples are clustered with the unknown samples, which aggravates the domain misalignment.  
Thus, it is unreliable to complete the domain alignment without any prior knowledge about the distribution of unknown samples.

Moreover, the biased classifier can produce overconfident predictions for unknown samples. Most UniDA methods employ one or more classifiers which produce a confidence for each target sample to determine whether it belongs to a particular known class seen in the source domain or the unknown class. Since they usually train their classifiers with the supervised source samples, the less-discriminative embeddings and the labeled objective in the source domain can lead the whole model to be biased towards the common classes of the target domain. This results in overconfident predictions of many samples belonging to the unknown class. In addition, as mentioned by Chen \textit{et al.}~\cite{chen2022geometric}, the class competition nature may also cause the model to generate overconfident predictions for unknown instances. To handle this issue, some recent approaches applied extra components to help classify the unknown samples. For instance, Fu \textit{et al.}~\cite{fu2020learning} employed multiple classifiers to detect the unknown target samples by a mixture of uncertainties. Saito \textit{et al.}~\cite{saito2021ovanet} proposed to use a one-vs-all classifier to distinguish the unknown samples and Chen~\textit{et al.}~\cite{chen2022geometric} extended the softmax-based classifier to produce an energy-based uncertainty for determining the unknown samples.

To address the above two issues, we propose a novel uncertainty-guided UniDA framework to reduce the influence of the domain misalignment and balance the confidences of known and unknown samples. First of all, without relying on the predictions output by the classifier, we introduce an empirical estimation of the posterior probability for a target sample being `unknown' through its neighborhood information in the source domain. Meanwhile, we prove that the proposed estimation is theoretically reliable. The estimation of the posterior probability reveals that the consistency between the labels of neighbors searched from the source domain and the distance between the target sample and its $k$-nearest neighbors are two keys to distinguish the known and unknown samples. Then, based on these two factors, we propose a novel neighbors searching scheme in a linear subspace with a $\delta$-filter to estimate the uncertainty of each target sample, which is employed to distinguish the known and unknown samples.
Firstly, to better discover unknown samples through the label-consistency of a target sample's neighbors, we project the features of source and target samples into a linear feature subspace to reduce the influence of the domain misalignment and improve the reliability of neighbors. As illustrated in the rightmost part of Fig.~\ref{intro}(b), projecting features in the original representational space into the linear subspace can reduce the correlation between all pairs of samples, which can make the unknown samples move away from the edges of the source clusters, and consequently the consistency of the labels of neighbors for an unknown sample decreases. 
Secondly, since the distance between a known sample and the centroid of a source class would not be significantly different from that between an unknown sample with the centroid, it is hard to find an optimal threshold to filter the discovered known samples through the $k$-nearest neighbor distance. Therefore, we propose to estimate the difference of the dispersions of two vector sets respectively. One set contains the target sample and its neighbors belonging to the same class, and another set consists of those neighbors and a randomly selected sample from the same source class.  The $\delta$-filter can well estimate that if a target sample is compact enough with most of its neighbors belonging to the same class.

For the second challenging problem, the classifier training on supervised source samples can be biased to the source classes, which can lead to the  inconsistency between intra-class variances of the source classes and the unknown target class. Thus, it easily produces overconfident predictions for the unknown samples.
To deal with that issue, we propose a novel uncertainty-guided margin loss (UGM) to encourage the intra-class variances of the source classes similar to that of the discovered unknown samples by an uncertainty adaptive margin mechanism. To avoid setting the margin term manually and better represent the intra-class variance of the unknown class, the margin term is produced based on the confidence level of the unknown samples automatically.

In summary, the contributions of this paper are thus fourfold:
\begin{itemize}
    \item We introduce an empirical estimation of the posterior probability for a target sample belonging to the unknown class which fully exploits the distribution of target samples in the latent space and theoretically prove the reliability of the proposed empirical estimation.

    \item Based on the estimation of the posterior probability, we propose a novel neighbors searching scheme in a linear subspace with a $\delta$-filter where features in the linear subspace can reduce the misalignment between source and unknown samples, and the $\delta$-filter can determine if a target sample is compact enough with respect to its neighbors.
    
    \item We present a novel uncertainty-guided margin loss to reduce the gap between the intra-class variances of the source classes and the unknown class which can balance the predictions of known samples and that of unknown samples.
    \item We perform experiments under various benchmarks. The results demonstrate that our method can significantly outperform baseline methods and achieve state-of-the-art performance.

\end{itemize}

\section{Related Work}
We briefly review recent methods which handle the UDA problems with non-identical label sets including PDA, ODA and UniDA in this section. In addition, we also briefly review a related problem named Out-of-Distribution detection to demonstrate the inspiration to our work.

\subsection{Partial-set Domain Adaptation}
In PDA setting, the label set of target domain is much smaller than and contained by that of the source domain.
Recently, many existing methods\cite{cao2018san,cao2018partial,zhang2018importance,cao2019learning,liang2020balanced,wang2022manifold} have been investigated to deal with the problem in PDA. 
Cao \textit{et al.}~\cite{cao2018san} solved this problem through a selective adversarial network (SAN). SAN simultaneously reduced the negative transfer and promoted positive transfer to align the distributions of samples from two domains in a fine-grained manner. Zhang \textit{et al.}~\cite{zhang2018importance} defined the domain similarities from a domain discriminator to identify common samples and applied a weighting scheme based on such similarities for the adversarial training. 
To better estimate the transferability of source samples, Cao \textit{et al.}~\cite{cao2019learning} proposed a progressive weighting operation. Liang \textit{et al.}~\cite{liang2020balanced} introduced a balanced
adversarial alignment to avoid the negative knowledge transfer and adaptive uncertainty suppression to reduce the uncertainty propagation.

\subsection{Open-set Domain Adaptation}

Compared to the PDA, ODA, firstly introduced by Busto \textit{et al.}~\cite{panareda2017open}, concerns the opposite scenario.  It assumes that some classes in the target domain are private and unseen to the source domain.
To address this challenging problem, Busto \textit{et al.}~\cite{panareda2017open}  introduced the Assign-and-Transform-Iteratively (ATI) algorithm to find the unknown samples.
Recently, one of the most popular strategies~\cite{liu2019separate,feng2019attract,gao2024survey,wang2021zeroth} for aligning the two domains in ODA is applying the domain discriminator to identify common samples across domains and exclude the unknown samples. Saito~\textit{et~al.}~\cite{saito2018open} proposed an adversarial learning framework to obtain a boundary between source and unknown samples whereas the feature generator was trained to locate the unknown samples far from the boundary. Bucci \textit{et al.}~\cite{bucci2020effectiveness} employed self-supervised learning to separate the known and unknown samples and complete the domain alignment.

\subsection{Universal Domain Adaptation}
UniDA, which is firstly introduced by You \textit{et al.} \cite{you2019universal}, concerns about the most general setting in UDA which is a more challenging problem than PDA and ODA, since the prior knowledge about the overlap of label sets between the two domains is unknown. You \textit{et al.} also proposed to evaluate the transferability of samples through a universal adaptation network (UAN) which estimated the uncertainty of target samples and domain similarity. However, measurements in \cite{you2019universal} are not robust and discriminative enough. Then, Fu \textit{et al.}~\cite{fu2020learning} proposed another transferability measure, called Calibrated Multiple Uncertainties (CMU). They evaluated the transferability and quantified the inclination of a target sample to the common classes by a mixture of uncertainties. Li \textit{et al.}~\cite{li2021DCC} introduced Domain Consensus Clustering (DCC)  to exploit the domain consensus knowledge for discovering discriminative clusters of target samples, which separated the unknown samples from the common ones. OVANet \cite{saito2021ovanet}, proposed by Saito \textit{et al.}, trained a one-vs-all classifier using labeled source samples for each source class to classify the known/unknown samples, and they adapted the open-set classifier to the target domain to classify the common ones. Recently, Chen \textit{et  al.}~\cite{chen2022geometric} proposed a geometric anchor-guided adversarial and contrastive learning framework with uncertainty modeling and achieve the  state-of-the-art (SOTA) by exploring a new neighbors clustering method to complete the domain alignment, and extend the traditional softmax-based classifier to the energy-based classifier. However, all recent methods do not consider that adapting the domain misalignment between two domains is dangerous since we do not have any knowledge about the source private classes and the unknown target samples. Especially, they could not perform well in the scenario of the unknown set being large.

\subsection{Out-of-Distribution Detection}
The problem of detecting outliers and anomalies in the data, which named as out-of-Distribution (OOD) detection, has been extensively studied. Since we should discover the outliers of the target domain in UniDA, OOD detection is closely related to our method. OOD detection has been greatly studied both in the supervised \cite{golan2018deep} and unsupervised \cite{wang2019effective} settings. To get some inspirations, we mainly focus on the recent deep learning based approaches with unsupervised settings. These methods either estimated the distribution of ID (i.e. In-Distribution) samples \cite{lee2018simple,pidhorskyi2018generative,liu2024graph} or used a distance metric between the test samples and ID samples to detect OOD samples \cite{liang2017enhancing,hendrycks2016baseline,chen2023machine}. Firstly, many of the existing  approaches employed the OOD datasets during training \cite{yu2019unsupervised,sun2022out,hendrycks2019using,gong2024clinical} or validation steps \cite{liang2017enhancing,lee2018simple,vyas2018out,pidhorskyi2018generative,lee2017training,ren2019likelihood,zhou2023interpretable}.
For instance, in \cite{yu2019unsupervised}, the network was fine-tuned during the training to decrease the inter-sample affinity between ID and OOD distributions. Other interesting methods, such as  \cite{lee2018simple,liang2017enhancing,vyas2018out,wang2019nonconvex}, applied a perturbation on each sample at test time to exploit the robustness of their network in detecting ID samples. However, they used some of the OOD samples to fine-tune the perturbation parameters. Moreover, 
methods that relied on generative models or auto-encoders, such as Pidhorskyi \textit{et al.} \cite{pidhorskyi2018generative}, also required hyper-parameter tuning for loss terms, regularization terms, and latent space size. The authors in \cite{shalev2018out} proposed to use extra supervision to construct a better latent space and to detect OOD samples with high accuracy through multiple semantic dense representations. Although having access to extra information certainly helped boost performance, it could be argued that OOD detectors should be completely agnostic of the unknown distributions, which was a more realistic scenario in the wild. Only a few approaches, such as \cite{hendrycks2016baseline,hsu2020generalized,liu2019large,neal2018open,yoshihashi2019classification,wang2024survey}, did not require the OOD samples neither during training nor validation. For instance, Hendricks \textit{et al.} and Gimpel \textit{et al.} \cite{hendrycks2016baseline} showed how the softmax layer can be used to detect OOD samples, when its prediction score is below a threshold. In  \cite{yoshihashi2019classification}, the authors relied on reconstructing the samples to produce a discriminative feature space. However, methods that relied on either reconstruction or generation \cite{neal2018open,yoshihashi2019classification,pidhorskyi2018generative,liu2024automatic}  did not perform well in scenarios where sample generation or reconstruction was more difficult, such as large-scale datasets. 
Although many methods in OOD detection are instructive for discovering the unknown samples in UniDA, we should also complete the domain alignment which is another challenging problem.




\begin{figure*}[t]
\centering
\includegraphics[width=1\linewidth]{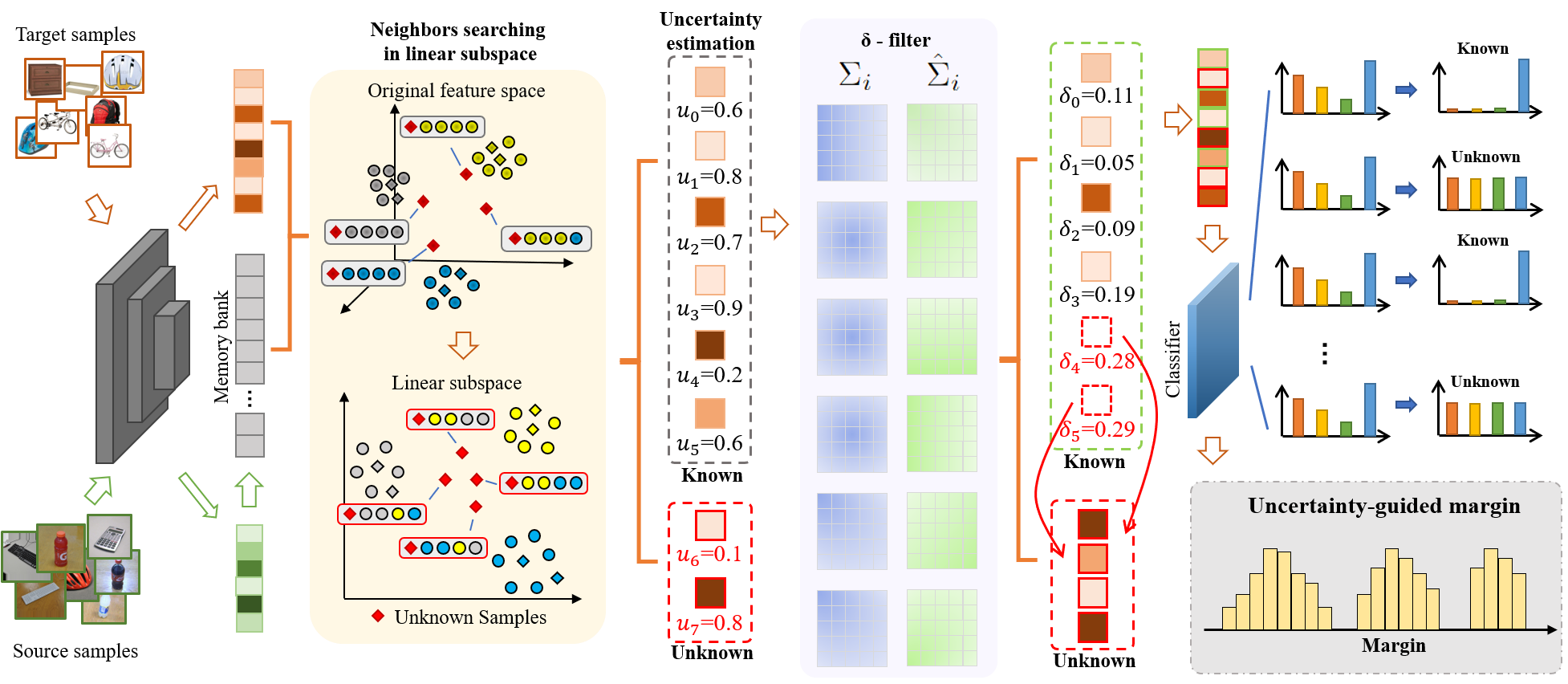}
\caption{The overall workflow of the proposed UniDA framework. By projecting the features extracted from the samples of both domains into a linear subspace, we estimate their uncertainty based on the neighbors searching and find the known/unknown samples. We refine the label of the known samples with the $\delta$-filter and  send all the target samples into the classifiers with pseudo labels. Based on the prediction of the unknown samples, we can compute the uncertainty-guided margin loss which balances the intra-class variances of both domains and leads to a reliable classification.
}
\label{fig:frame}
\end{figure*}

\section{Method}
In UniDA, we have a labeled set of source domain $D^s = \{({\mathbf{z}^s_i}, y^s_i)\}$ and an unlabeled set of target domain ${D}^t = \left\{{{\mathbf{z}^t_i}}\right\}^{N_t}_{i=1}$. With defining the source and target label sets as $Y^s$ and $Y^t$ respectively, we denote $Y^{com} = Y^s \cap Y^t$ as the common label set and $Y^{unk} = \complement_{Y^t}{Y^{com}}$ as the unknown label set, where $\complement_{A}{B}$ means the complement of set $B$ in set $A$. With assuming that the source label set $Y^s$ containing $C$ classes, we denote the unknown class as $C+1$ for convenience. The method aims to train an optimal classifier $\mathcal{C}: {Z}^t \rightarrow Y^s$ on both domains and categorize a target sample into  one of the $C+1$ classes. In this section, we introduce an empirical estimation of the probability of a target sample belonging to the unknown class. Then, we elaborate the major components of our method {in} the training process  which sufficiently avoids the influence of the domain misalignment and  balances the confidences of known and unknown samples. 

\subsection{Empirical Estimation of Unknown Samples }
In this section, we introduce an empirical estimation of the posterior probabilities for unknown samples by leveraging the neighborhood information of a target data which can distinguish most of the unknown samples more reliably. Theoretically we prove that our empirical estimation of the posterior probability $p(y\!=\!C\!+\!1|\mathbf{z})$  is reliable. Unlike most existing methods \cite{you2019universal,saito2020universal,fu2020learning,saito2021ovanet} relying on the posterior probability of a softmax-based classifier, we focus on the how the target samples 
are distributed in the latent representational space and the relationship between samples from two domains. 

\newtheorem{theorem}{Theorem}
\newtheorem{lemma}[theorem]{Lemma}
\newtheorem{proposition}[theorem]{Proposition}

\newtheorem{corollary}[theorem]{Corollary}
\newtheorem*{proof}{Proof}



\begin{proposition}\label{theorem}
 \textit{With the  feature set of samples from two domains $Z^s$, $Z^t$ and a target feature $\mathbf{z} \in Z^t$, denoting     $\hat{p}_{C\!+\!1}(\mathbf{z};k) = c_1\mathbbm{l}\{\max_{i=0,\dots,C}{\hat{p}_i(\mathbf{z};k,k_i\mid y=i)}\leq \beta\}$ 
 where $k$ and $k_i$ are the number of neighbors and the number of neighbors belonging to class $i$, respectively, we have:}
 
\textit{If}
\begin{equation}
    c_0\frac{k_{max}}{k(r_k)^{m-1}}\leq \beta
\label{assume1}
\end{equation}

\textit{Then,}
\begin{equation}
\label{leq1}
    \mathbbm{l}\{p\left(y\!=\!C\!+\!1  \!\mid\! \mathbf{z}\right)<\gamma\}=\mathbbm{l}\{(r_k)^{m-1}<\frac{\epsilon c_0 \gamma}{(1-\gamma)(1-\epsilon)}\}
\end{equation}
\textit{where  $k_{max}=max_{i=0,\dots,C}(k_i)$, $\gamma\in [0,1]$, $r_k$ is the $k$-nearest neighbor distance, $c_{0,1}$ and $\epsilon$ are non-zero constants. All samples in the feature space are {normalized} (i.e. $\mathbf{z}_i = \frac{\mathbf{z}_i}{|\mathbf{z}_i|}$).}
 \end{proposition}
  
\begin{proof}
We provide the proof sketch to show our key ideas which revolves around performing the empirical estimation of  $p(y\!=\!C\!+\!1|\mathbf{z})$.

First, since we have no idea which known class is private for the source domain, we denote:
\begin{equation}
\label{1-kn}
    p(y\!=\!C\!+\!1|\mathbf{z}) = 1 - p(y\!\in\!Y^s|\mathbf{z}) 
\end{equation}
and we estimate the posterior distribution of $\mathbf{z}$ belonging to one of source classes which is easier.

By Bayes' rule, the probability of $\mathbf{z}$ belonging to one of the source classes can be found as: 
\begin{equation}
\label{bay}
\begin{aligned}
&p\left(y\in Y^s  \!\mid\! \mathbf{z}\right) =\frac{ p\left(\mathbf{z} \!\mid\! y\in Y^s\right) \cdot p\left(y\in Y^s\right)}{p\left(\mathbf{z}\right)} \\
&=\frac{ {\sum\limits^{C}_{i=0} p_{i}\left(\mathbf{z}\right) \cdot\sum\limits^{C}_{j=0} p\left(y=j\right)}}{\sum\limits^{C}_{i=0} p_{i}\!\left(\mathbf{z}\right)\! \cdot\!\sum\limits^{C}_{j=0} p\left(y=j\right)+p_{{C\!+\!1}}\!\left(\mathbf{z}\right) \!\cdot\! p\left(y=C \!+\!1\right)}
\end{aligned}
\end{equation}

\end{proof}
Then, the estimation of $p(y\in Y^{s} \mid \mathbf{z})$ reduces to deriving the estimations of probability density functions $p_{0}(\mathbf{z}),\dots,p_{C+1}(\mathbf{z})$.

\begin{lemma}\label{lemma}
\textit{With  $k$, $k_i$ and $r_k$ defined in Proposition \ref{theorem}, we can estimate the probability density function $p_i(\mathbf{z})$ as:}

\begin{equation}
\label{pzkki}
    \hat{p}_i(\mathbf{z};k,k_i)=c_0\frac{k_i}{k(r_k)^{m-1}}.
\end{equation}

\end{lemma}

\begin{proof}
Since ${\mathbf{z}\in\mathbb{R}^m}$ and all features are normalized where $||\mathbf{z}||\!=\!1$, all data points are located on the surface of an $m$-dimensional unit sphere. We set $U_r(\mathbf{z}) = \left\{ || \mathbf{z}^{\prime} - \mathbf{z} ||_{2} \leq r \right\}\cap\{\mathbf{z}^{\prime}\!\in\! Z^s\} $, which is a set of data points from source domain on the unit hyper-sphere centered on $\mathbf{z}$ with a radius $r$. Assuming the density probability functions satisfy Lebesgue's differentiation theorem, the probability density function can be attained by:
\begin{equation}
    p_{i}(\mathbf{z}) = \lim_{r\rightarrow 0}\frac{p(\mathbf{z}^{\prime}\in U_r(\mathbf{z},r)| {y={j}})}{|U_r(\mathbf{z})|}
\end{equation}
Since $r_k$ dnotes the Euclidean distance between the center $\mathbf{z}$ and its $k$-$th$ nearest neighbor, we get:
\begin{equation}
\begin{aligned}
\hat{p}_{i}(\mathbf{z};k) 
&= \frac{p(\mathbf{z}^{\prime}\in U_{r_k}(\mathbf{z})|\mathbf{z}^{\prime}\in Z^s_i)}{|U_{r_k}(\mathbf{z})|}.
\end{aligned}
\end{equation}

Denoting $B_i$ as the smallest sphere containing $Z^s_i=\{\mathbf{z}_1^{\prime}\dots\mathbf{z}_{l}^{\prime}\}$, where $Z^s_i$ is the set of all source samples belonging to class $i$. We can then assume that:
\begin{equation}
    \forall i,j\in\{0,\dots,C\}; B_i\cap B_j=\emptyset
\end{equation}
Then, we have:

\begin{equation}
    \hat{p}(\mathbf{z}^{\prime}\in U_{r_k}(\mathbf{z})|\mathbf{z}^{\prime}\in Z^s_i) = \frac{|U_{r_k}(\mathbf{z})\cap B_i|}{|U_{r_k}(\mathbf{z})|}.
\end{equation}

We assume the number of neighbors is big enough. Then, we have the estimation $|U_{r_k}(\mathbf{z})\cap B_i| = c_0\frac{k_i}{k}$ where $c_0$ is a constant. Then, the approximation of $\hat{p}_i(\mathbf{z};k)$ can be attained by:
\begin{equation}
\label{eq10}
    \hat{p}_i(\mathbf{z};k,k_i)=c_0\frac{k_i}{k(r_k)^{m-1}}
\end{equation}
where $k_i$ is the number of samples belonging to $U_{r_k}$ and $B_i$.
\end{proof}

Another challenge of estimating $p(y\!\in\!Y^s|\mathbf{z})$ is computing $p_{C\!+\!1}$ since we do not have any prior knowledge about unknown samples. The only knowledge we have is that samples not belonging to all the source classes are unknown samples. Thus, we obtain:


\begin{equation}
    \hat{p}_{C\!+\!1}(\mathbf{z};k) = c_1\mathbbm{l}\{\max_{i=0,\dots,C}{\hat{p}_i(\mathbf{z};k,k_i)}\leq \beta\}
\label{asspc}
\end{equation}
where $\beta$ is a constant chosen to satisfy the equation.

\begin{lemma}
\textit{With the assumption  that $\hat{p}_{C\!+\!1}(\mathbf{z};k)$ satisfies Eq. (\ref{asspc}), we can infer Eq. (\ref{leq1}) with the restriction Eq. (\ref{assume1})}.
\end{lemma}

\begin{proof}
 According to Eqs. (\ref{1-kn}), (\ref{bay}) and (\ref{pzkki}),
denoting $\sum^C_{i=0}p(y\!=\!j) = \varepsilon$ and $k_{max}=max_{i=0,\dots,C}(k_i)$, for a real number $\gamma \in [0,1]$ we have:

If
\begin{equation}
    c_0\frac{k_{max}}{k(r_k)^{m-1}}\leq \beta
\end{equation}

Then,


\begin{equation}
\label{leq}
\begin{aligned}
&\quad\,\,\mathbbm{l}\{p\left(y\!=\!C\!+\!1  \!\mid\! \mathbf{z}\right)<\gamma\}\\
&=\mathbbm{l}\{1-p\left(y\in Y^s  \!\mid\! \mathbf{z}\right))>\gamma\}\\ 
&=\mathbbm{l}\{\frac{(1-\epsilon)  \hat{p}_{C+1}(\mathbf{z;}k)}{\epsilon \sum^C_{i=0}\hat{p}_{i}(\mathbf{z;}k) + (1-\epsilon)  \hat{p}_{C+1}(\mathbf{z;}k)}>\gamma\}\\
&=\mathbbm{l}\{\frac{(1-\epsilon)}{\frac{\epsilon c_0}{(r_k)^{m-1}} + (1-\epsilon)}>\gamma\}\\
&=\mathbbm{l}\{(r_k)^{m-1}<\frac{\epsilon c_0 \gamma}{(1-\gamma)(1-\epsilon)}\}\\
\end{aligned}
\end{equation}
In addition, when $ c_0\frac{k_{max}}{k(r_k)^{m-1}}> \beta$, we have:
\begin{equation}
    p (y = C +1 | \mathbf{z})= 0.
\end{equation}
\end{proof}

Notably, in the ODA problem \cite{saito2018open,saito2018maximum} where the source label set is contained in the target label set, we can get the equation:
\begin{equation}
\label{s=com}
    p(y\!\in\! Y^s|\mathbf{z}) = p(y\!\in\! Y^{com}|\mathbf{z})
\end{equation}
Next, we can reliably estimate the probability of a target sample belonging to one of the common classes  based on Eqs. (\ref{bay}) and (\ref{pzkki}).  However, in the OPDA problem \cite{saito2020universal,bucci2020effectiveness}, Eq. (\ref{s=com}) would not work any more because of the existence of the private source classes. 

Then, from Proposition \ref{theorem}, when $k_{max}$ satisfies Eq. (\ref{assume1}), the upper boundary of the probability $p(y=C+1\mid\mathbf{z})$ is positively  correlated with the $k$-nearest neighbor distance $r_k$, i.e., when $k_{max}$ and $r_k$ are big enough,  bigger $r_k$ means the bigger probability of a target sample being unknown.

\begin{corollary}
\label{corollary}
\textit{In the UniDA problem, a known target sample {$\mathbf{z}$  should satisfy the following} conditions:
\begin{itemize}
    \item The neighbors of $\mathbf{z}$ should mostly belong to one of the {common classes}.
    \item  $\mathbf{z}$ should be close enough to its neighbors.
\end{itemize}}
\end{corollary}

Based on the conditions mentioned in Corollary~\ref{corollary}, the reliable neighbors searching scheme is necessary because the domain misalignment can cause the mismatch between the target samples and the source samples.



\begin{figure}[t]
  \centering
   \includegraphics[width=0.95\linewidth]{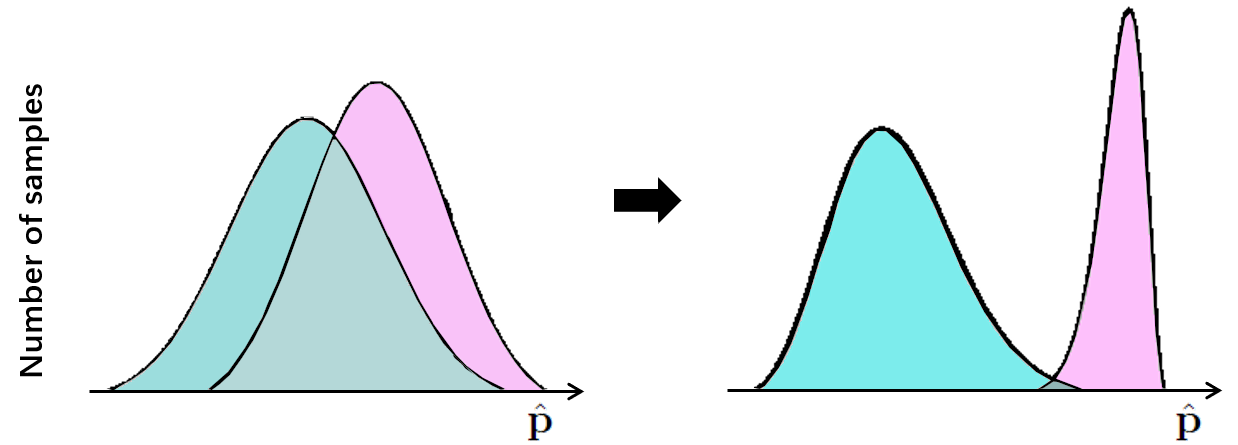}
   \vspace{-1mm}
   \caption{Illustration for the effect of NSLS. Extracting the principal linear subspace can reduce the overlap of $\hat{p}$ of known target samples and unknown target samples.}
   \label{nsls}
\end{figure}

\subsection{Discovering Unknown Samples Based on the Uncertainty Estimation}
To improve the neighbors searching scheme based on $k$-NN algorithm and discover unknown samples reliably, we propose an uncertainty estimation method to discover unknown samples. It is based on the neighbor searching in a linear subspace and a $\delta$-filter to justify if a target sample can satisfy conditions in Corollary \ref{corollary} respectively.

\subsubsection{Neighbors searching in linear subspace (NSLS)}
Since some known samples may be distributed far away from the centers of the source classes, and unknown samples may be distributed in the edges of the source clusters due to the domain discrepancy,
clustering target samples with their nearest neighbors or nearest prototypes is dangerous in the original feature space. 
Therefore,  we propose to find a reliable linear subspace to deal with the above problems and
improve the accuracy of discovering the unknown samples based on the neighborhood information searched from source domain. Specifically, given $  Z^s=[\mathbf{z}^s_0,\mathbf{z}^s_1, \cdots, \mathbf{z}^s_{N_s}]^{\mathsf{T}}\in \mathbb{R}^{N_s\times m}$ and $ \mathcal{B}=[\mathbf{z}^t_0,\mathbf{z}^t_1,\cdots,\mathbf{z}^t_b]^{\mathsf{T}} \in \mathbb{R}^{b\times m}$, where $m$  {means} the dimension of $\mathbf{z}_i$, $Z^s$ and $\mathcal{B}$ are {sets} of all source samples and target samples in a mini-batch respectively, we denote the original feature set $Z=[\mathbf{z}_0^s,\mathbf{z}^s_1, \cdots, \mathbf{z}^s_{N_s},\mathbf{z}^t_0,\mathbf{z}^t_1,\cdots,\mathbf{z}^t_b] \in \mathbb{R}^{n\times m}$ where $n=N_s+b$.

Then, we have
\begin{equation}
    Z_{sub} = Z_{}P, \,\, P\in \mathbb{R}^{m \times p},Z_{sub}\in \mathbb{R}^{n \times p}
\label{dim_reduce}
\end{equation}
where $P$ is a transformation matrix mapping the features with $m$ dimensions to reduced features with $p$ dimensions.

To get the $Z_{sub}$, 
we propose to analyze the covariance matrix of $Z$. After centralizing $Z$, the covariance matrix $A$ can be defined as:
\begin{equation}
    A = \frac{1}{n}ZZ^{\mathsf{T}} =[cov(\mathbf{z}_i\mathbf{z}_j)]^{n\times n}_{i,j=0,\dots,n}.
\end{equation}

Inspired by Wang \textit{et al.} \cite{wang2021regularizing},  the covariance matrix captures the feature distribution of the training data, and contains rich information of potential semantic differences. We propose to decompose the covariance matrix $A$ to find the dimensions which can best represent the semantic difference of each source class and cut off the other dimensions. After the reduction of dimensions, since the unknown target samples do not share the common features with the known source classes, the distances from an unknown sample to each of the source classes can be averaged which can lower $k_{max}$ in Eq. (\ref{assume1}). Specifically, we leverage the singular value decomposition (SVD) method to decompose the covariance matrix $A$:
\begin{equation}
    A = {U_{m\times m}} {\Sigma_{m\times m}} {V_{m\times m}}.
\end{equation}

Then, we can get the transformation matrix $P$ from $V$:
\begin{equation}
    P=V_{{m \times p}}^{\mathsf{T}}.
\end{equation}

\begin{figure*}[t]
  \centering
   \includegraphics[width=0.9\linewidth]{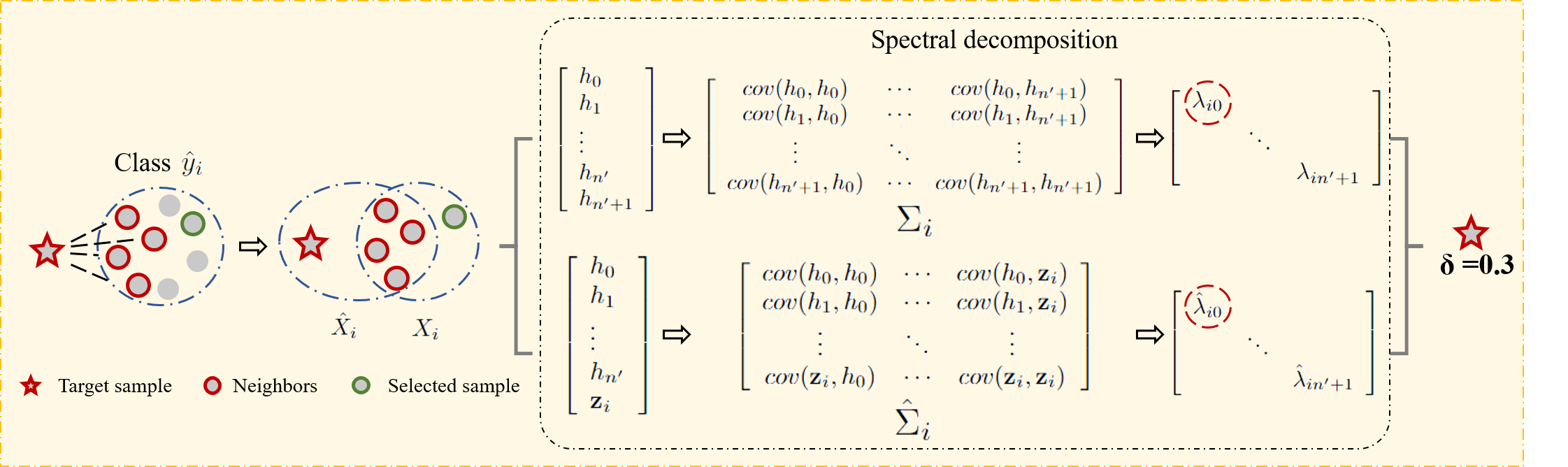}
   \vspace{-1mm}
   \caption{Computation of $\delta$: the target sample  should be compact enough with its neighbors compared to the compactness of the source class. The terms $\lambda_{i0}$, $\lambda_{in^{\prime}+1}$, $\hat{\lambda}_{i0}$, $\hat{\lambda}_{in^{\prime}+1}$  represent the singular values.}
   \label{filter}
\end{figure*}



By projecting the original features to the linear subspace, we propose to search for the neighbors of each of the target samples in the subspace. 
Particularly, we firstly employ a memory bank $M$ to store all the features of the source samples:
\begin{equation}
    M = [\textbf{m}_0,\dots,\textbf{m}_n] 
\end{equation}
with a momentum scheme to update the memory bank:
\begin{equation}
\label{update}
     \textbf{m}_i = \alpha \textbf{m}_i + (1-\alpha)\mathbf{z}_i.
\end{equation}
Then, we project the set $Z=[\textbf{m}_0,\dots,\textbf{m}_{N_s},\mathbf{z}^t_0,\mathbf{z}^t_1,\cdots,\mathbf{z}^t_b]$ to $Z_{sub}$. We search for the neighbors $\mathcal{N}$ for each target sample $\mathbf{z}_i$ in $Z_{sub}$ and propose an uncertainty score of $\mathbf{z}_i$ according to Eq.(\ref{eq10}):
\begin{equation}
\label{est}
    u(\mathbf{z}_i) = \max_{[i = 0,\dots,Y]}(|\{ \mathbf{z}^{\prime} \in \mathcal{N}^k | y^{\prime}=i\}|)
\end{equation}
where $k$ is the number of searched neighbors.
Then, the discovered unknown samples are denoted as:
\begin{equation}
    \hat{Z}^{unk}=\{ \mathbf{z}_i\mid u(\mathbf{z}_i) \leq \tau\}
\end{equation}
where $\tau$ is set manually. Similarly, the known samples are denoted as:
\begin{equation}
    \hat{Z}^{k}=\{ \mathbf{z}_i\mid u(\mathbf{z}_i) > \tau\}
\label{kset}
\end{equation}  
and $\mathbf{z}_i\in \hat{Z}^{k}$ has pseudo label $\hat{y}_i$ which is defined as:
\begin{equation}
    \hat{y}_i = {argmax} _ {[i = 0,\dots,Y]}(|\{ \mathbf{z}^{\prime} \in \mathcal{N}^k | y^{\prime}=i\}|).
\end{equation}

\subsubsection{$\delta$-filter  for discovering unknown samples }  
As mentioned in the second condition of Corollary \ref{corollary}, the target sample should be close enough to its neighbors. However, `nearest' neighbors is not equivalent to `most compact' neighbors. For instance, for an unknown sample, which is far away from all source classes, its `nearest' neighbors can also mostly belong to the same class which means that there are some noisy data in $\hat{Z}^k$.
Moreover, since known ones might also distribute on the edges of clusters of the source classes, the distance from a centroid of a source class to a known sample might not be significantly different from the distance between the centroid and an unknown sample. Thus, it is hard to
filter noisy known samples from $\hat{Z}^k$ through the $k$-nearest neighbor distance. Therefore we introduce a dynamical $\delta$-filter scheme to remove the noisy data.


To be specific, as shown in Fig. \ref{filter},  compared to the compactness of the class $\hat{y}_i$,  a credible known sample $\mathbf{z}_i\in \hat{Z}^k$ with its neighbors  belonging to class $\hat{y}_i$ should be compact enough. To estimate the compactness of $\mathbf{z}_i$ and its neighbors, we apply the spectral decomposition on the covariance matrices of  these vectors ( i.e., $\mathbf{z}_i$ and its neighbors belonging to class $\hat{y}_i$) to get the maximum eigenvalues of the covariance matrices which can represent the dispersion of the vectors. 
Moreover, to get a comparable estimation of the dispersion of vectors in class $\hat{y}_i$, we randomly select a sample from the source class $\hat{y}_i$ which is not a neighbor of $\mathbf{z}_i$ and compute the maximum eigenvalue of the covariance matrix of the selected vector and the neighbors belonging to class $\hat{y}_i$. We leverage the difference of the two eigenvalues to determine whether $\mathbf{z}_i$ is noisy.

In detail, for $\mathbf{z}_i\in \hat{Z}^k$, we denote $\mathcal{N}^{\prime}_{i}=\{h_0,h_1,\dots,h_{n^{\prime}}\}$ as the neighbors of $\mathbf{z}_i$ where samples in $\mathcal{N}^{\prime}_{i}$ belong to class $\hat{y}_i$ and $n^{\prime}=|\mathcal{N}^{\prime}_{i}|>\tau$. Next, we can denote two zero-mean matrices defined as:
\begin{equation}
\begin{aligned}
&X_i=centralize([h_0,h_1,\dots,h_{n^{\prime}},h_{n^{\prime}+1}]),\\ 
&\hat{X}_i=centralize([h_0,h_1,\dots,h_{n^{\prime}},\mathbf{z}_i])
\label{mat}
\end{aligned}
\end{equation}
where $h_{n^{\prime}+1}$ means the selected source sample. Then, we can get the covariance matrices of both sets:
\begin{equation}
    \Sigma_{i}=\frac{1}{n^{\prime}+1}X_i X_i^{\mathsf{T}},\quad \hat{\Sigma}_i=\frac{1}{n^{\prime}+1}\hat{X}_i \hat{X}_i^{\mathsf{T}}.
\label{sig}
\end{equation}
We leverage spectral decomposition on both covariance matrices to get maximum eigenvalues $\lambda_i$ and $\hat{\lambda}_i$ of $\Sigma_i$ and $\hat{\Sigma}_i$ respectively. Finally, we define the difference value $\delta_i$ as:
\begin{equation}
    \delta_i=|\lambda_i-\hat{\lambda}_i|.
\label{delta}
\end{equation} 
When $\delta_i$ is bigger, it means that including $\mathbf{z}_i$ has a significantly bad influence on the description of the vector set and $\mathbf{z}_i$ should be filtered out from $\hat{Z}^k$. 
Experimentally, we assume that a clean known target sample should satisfy $\delta_i \leq 0.2\lambda_i$. 

\subsection{Learning}

With discovered unknown samples, we aim to train the classifier to categorize a target sample into one of the source classes with high confidence and distinguish target samples belonging to the unknown class via the entropy of the output.
Thus, the training of classifier  involves a trade-off: maximizing classification performance on $Z^{com}$ and preventing overconfident predictions on $Z^{unk}$. Denoting $W=[W_{0}^{\mathsf{T}},W_{1}^{\mathsf{T}},\dots,W_{Y^s}^{\mathsf{T}}]$ as the weights of a linear classifier, a traditional method\cite{saito2021ovanet,cao2019learning,li2021DCC} is to improve the classification performance by training the classifier in the source domain with a cross-entropy loss:
\begin{equation}
\label{celoss}
\mathcal{L}_{s}=\frac{1}{n} \sum_{\mathbf{z}_{i}^s \in Z^s}-\log \frac{e^{W_{y_{i}}^{\mathsf{T}}  \mathbf{z}_{i}}}{e^{W_{y_{i}}^{\mathsf{T}}  \mathbf{z}_{i}}+\sum_{j \neq i} e^{W_{{j}}^{\mathsf{T}}  \mathbf{z}_{i}}}
\end{equation}
where $ W_{y_{i}}^{\mathsf{T}} $ is the weight of class $y_i$ which is the ground truth of $\mathbf{z}_i^s$.

However, this method can usually lead to the overconfident predictions on unknown samples~\cite{liang2020balanced}. To deal with the problem, Cao \textit{et al.}\cite{cao2019learning} aggregated multiple complementary uncertainty measures, OVANet \cite{saito2021ovanet} employed a one-vs-all classifier for  classifying unknown samples and GATE \cite{liang2020balanced} proposed an energy-based classifier which extended the traditional softmax-based classifier to improve the classification performance on unknown samples. In this work, without applying extra parameters,  we propose a more efficient method to train only one classifier via three losses.

\begin{table}[]
    \centering

\resizebox{0.95\textwidth}{!}{
\begin{tabular}{l}
\hline \multicolumn{1}{l}{{\bf Algorithm 1} Full algorithm of our method}\\

\hline {\bf Requirement:}\\
Source dataset ($Z^{s}$, $Y^{s}$), target dataset $Z^{t}$,\\
the number of neighbors $k$ and the threshold $\tau$.\\
{\bf Training:}\\
{\bf while} step $<$ max step do\\
$\quad$ {\bf if} step $= 0$ do\\
$\qquad\quad$ Extract all features from $Z^s$ and initialise  $M$\\
$\quad$ Sample batch $\mathcal{B}^s$ from ($Z^{s}$, ${Y}^{s}$) and batch $\mathcal{B}^t$ from ${Z}^{t}$ \\
$\quad$ Extract features from each of $\mathcal{B}^s$ and $\mathcal{B}^t$ \\

$\quad$ {\bf for} $\mathbf{z}^{s}_i \in \mathcal{B}^s$ do \\
$\qquad\quad$ Update $M$ by Eq. (\ref{update})\\
$\quad$ {\bf for} $\mathbf{z}^{t}_i \in \mathcal{B}^t$ do \\
$ \qquad\quad$ Retrieve the nearest neighbors $\mathcal{N}_i$ for $\mathbf{z}^{t}_i$ from $M$ \\
$ \qquad\quad$ Compute the uncertainty score $u(\mathbf{z}_i)$ as Eq. (\ref{est}) \\
$ \qquad\quad$ Compute $\sum_i$ and $\hat{\sum}_i$ by Eq. (\ref{sig})\\
$ \qquad\quad$ Decompose $\sum_i$ and $\hat{\sum}_i$ get $\lambda_i$ and $\hat{\lambda}_i$ \\
$ \qquad\quad$ Compute $\delta_i$ by Eq. (\ref{delta})\\

$ \qquad\quad$ {\bf if} $u(\mathbf{z}^t_i)<\tau$ and \textcolor{red}{$\delta\geq 0.2\lambda_i$} do\\
$\qquad\quad\qquad$ Append $\mathbf{z}^t_i$ into $\hat{Z}^{unk}$\\
$ \qquad\quad$ {\bf else} do\\

$\qquad\quad\qquad$ Append $\mathbf{z}^t_i$ into $\hat{Z}^{k}$\\
$\quad$ Compute $\mathcal{L}_{sup}$ based on Eq. (\ref{lsup})\\
$\quad$ Compute the margin $\mu$ based on Eq. (\ref{mu}) \\
$\quad$ Compute {$\mathcal{L}_{ugm}$} based on Eq. (\ref{margin loss})\\
$\quad$ Compute $\mathcal{L}_{unk}$ based on Eq. (\ref{lossunk})\\
$\quad$ Compute the overall loss $\mathcal{L}_{all}$\\ 
$\quad$ Update the model\\
\hline
\end{tabular}}

\end{table}

\subsubsection{Uncertainty-guided margin loss (UGM)}
Using the traditional CE-loss training on the source domain usually leads to a problem of imbalance between the predictions of the known/unknown target samples. Learning on source classes $Y^s$ is much faster than that on the unknown class $Y^{unk}$ due to the supervised objective which leads the whole model to be biased towards the common classes $Y^{com}$ in the target domain. As a result, the intra-class variance of the source domain can be much smaller than that of the target domain, including known target classes and the unknown class. The biased model can cause the overconfident predictions for many unknown samples.
Therefore, it is important to balance the intra-class variances of both domains. We propose a new uncertainty-guided margin loss (UGM) to achieve that. Specifically, at the beginning of the training, we enforce a larger margin to encourage a larger intra-class variance which is similar to the intra-class variance of discovered unknown samples. The margin $\mu$ goes down to zero close to the end of the training,
\begin{equation}
\label{margin loss}
\mathcal{L}_{ugm}=\frac{1}{n} \sum_{\mathbf{z}_{i} \in \mathcal{Z}_{l}}-\log \frac{e^{s(W_{y_{i}}^{\mathsf{T}}   \mathbf{z}_{i}+\alpha\mu)}}{e^{s(W_{y_{i}}^{\mathsf{T}}   \mathbf{z}_{i}+\alpha\mu)}+\sum_{j \neq y_i} e^{sW_{{j}}^{\mathsf{T}}   \mathbf{z}_{i}}}
\end{equation}
where the margin $\mu$ represents the intra-class variance of $\hat{Z}^{unk}$ which is defined as:
\begin{equation}
    \mu = \frac{1}{|\hat{Z}^{unk}|} \sum_{\mathbf{z}_j\in \hat{Z}^{unk}} \max(0,\max_{k=0,\dots,C} p(y = k| \mathbf{z}_j) - \frac{1}{2})
\label{mu}
\end{equation}
where $\mid\cdot\mid$ means the number of elements in a set.
We normalize the weights and inputs, i.e.,  $W_{y_{i}}^{\mathsf{T}} = \frac{W_{y_{i}}^{\mathsf{T}}}{|W_{y_{i}}^{\mathsf{T}}|}$ and $\mathbf{z}_i = \frac{\mathbf{z}_i}{|\mathbf{z}_i|}$. 

\subsubsection{Loss for unknown samples}

Since we use the entropy of the classifier output to distinguish the unknown samples, we need to lower the confidence of unknown samples belonging to $\hat{Z}^{unk}$. We employ the $\mathcal{L}_{unk}$ to smooth the posterior distribution for unknown inputs to increase the entropy:  
\begin{equation}
\label{lossunk}
    \mathcal{L}_{unk} = -\frac{1}{2|\hat{Z}^{unk}|}\sum_{\mathbf{z}_i\in \hat{Z}^{unk}}\sum_{i=0}^{Y} \log D(\mathbf{z}) - logY 
\end{equation}
where the inputs of Eq. (\ref{lossunk}) belong to the discovered unknown samples set $\hat{Z}^{unk}$.

\begin{table*}[t]
\caption{Comparison of main results on Office-31. Some results for previous methods are cited from OVANet \cite{saito2021ovanet} and GATE \cite{chen2022geometric}.}
\label{officeopda}

   \centering
   
   \resizebox{1\textwidth}{26mm}{
   
   \begin{tabular}{l|cccccc|c||cccccc|c}
   \hline \multirow{3}*{ Method }&\multicolumn{7}{c||}{ {\bf OPDA (H-score)}}&\multicolumn{7}{c}{ {\bf ODA (H-score)}}\\
   \cline{2-15} {} & \multicolumn{6}{c|}{ Office-31 $(10 / 10 / 11)$} & \multicolumn{1}{l||}{} & \multicolumn{6}{c|}{ Office-31 $(10 / 5 / 50)$} & \multicolumn{1}{l}{} \\
    & A2D & A2W & D2A & D2W & W2D & W2A & Avg& A2D & A2W & D2A & D2W & W2D & W2A & Avg \\
    \hline UAN \cite{you2019universal} & $59.7$ & $58.6$ & $60.1$ & $70.6$ & $71.4$ & $60.3$ & $63.5$ & $38.9$ & $46.8$ & $68.0$ & $68.8$ & $53.0$ & $54.9$ & $55.1$\\
    CMU \cite{fu2020learning} & $68.1$ & $67.3$ & $71.4$ & $79.3$ & $80.4$ & $72.2$ & $73.1$ & $52.6$ & $55.7$ & $76.5$ & $75.9$ & $64.7$ & $65.8$ & $65.2$\\
    DANCE \cite{saito2020universal} & $78.6$ & $71.5$ & $79.9$ & $91.4$ & $87.9$ & $72.2$ & $80.3$ & $84.9$ & $78.8$ & $79.1$ & $78.8$ & $88.9$ & $68.3$ & $79.8$ \\
    DCC \cite{li2021DCC} & $\mathbf{8 8 . 5}$ & $78.5$ & $70.2$ & $79.3$ & $88.6$ & $75.9$ & $80.2$ & $58.3$ & $54.8$ & $67.2$ & $89.4$ & $80.9$ & $85.3$ & $72.6$\\
    ROS \cite{bucci2020effectiveness} & $71.4$ & $71.3$ & $81.0$ & $94.6$ & $9 5 . 3$ & $79.2$ & $82.1$ & $82.4$ & $82.1$ & $77.9$ & $96.0$ & $99.7$ & $77.2$ & $85.9$\\
    USFDA \cite{kundu2020universal} & $85.5$ & $7 9 . 8$ & $8 3 . 2$ & $90.6$ & $88.7$ & $81.2$ & $84.8$ & $85.5$ & $7 9 . 8$ & $8 3 . 2$ & $90.6$ & $88.7$ & $81.2$ & $84.8$\\
    OVANet \cite{saito2021ovanet} & $85.8$ & $79.4$ & $80.1$ & $9 5 . 4$ & $94.3$ & $8 4 . 0$ & $8 6 . 5$ & $89.2$ & $\mathbf{89.0}$ & $86.4$ & $\mathbf{9 7 . 1}$ & $98.2$ & $\mathbf{8 8 . 1}$ & $91.3$  \\
    GATE \cite{chen2022geometric} & $87.7$ & $81.6$ & $\mathbf{84.1}$ & $9 4 . 8$ & $94.1$ & $8 3 . 4$ & $8 7 . 6$ & $88.4$ & $86.5$ & $84.2$ & $9 5 . 0$ & $96.7$ & $8 6 . 1$ & $8 9 . 5$ \\
    \hline
    \rowcolor{gray!30}
    Ours & $87.3$ & $\mathbf{83.5}$ & $82.2$ & $\mathbf{96.1}$ & $\mathbf{99.2}$ & $\mathbf{84.7}$ & $\mathbf{88.8}$ & $\mathbf{91.9}$ & $88.8$ & $\mathbf{86.9}$ & $94.8$ & $\mathbf{99.7}$ & $86.7$ & $\mathbf{91.5}$\\
   \hline
    \end{tabular}}

  \end{table*}

\subsubsection{Supervised contrastive loss on source domain}
Moreover, to reduce the influence of the domain misalignment, it is necessary to make each source class more compact and discriminative to enlarge the gap between two source classes which can improve the consistency of neighbors’ labels for a known target sample when it searches the nearest neighbors from 
the source domain. Thus, we employ a supervised contrastive learning loss~\cite{yu2019unsupervised} $\mathcal{L}_{sup}$ to make data points in the source domain more compact by pushing the samples from different classes apart while pulling the samples from the same class closer. $\mathcal{L}_{sup}$ is given by:
\begin{equation}
\label{lsup}
    \mathcal{L}_{sup}=\frac{1}{|\mathcal{B}^s|}\sum_{i=0}^{|\mathcal{B}^s|}\frac{\sum_{\mathbf{m}_j\in\mathcal{A}^{+}_i}\exp({<\mathbf{z}_i,\mathbf{m}_j>}/t)}{\sum_{\mathbf{m}_j\in \mathcal{A}^{-}_i\cup \mathcal{A}^{+}_i}\exp({<\mathbf{z}_i,\mathbf{m}_j>}/t)}
\end{equation}
where $\mathcal{A}^{+}_i$ and $\mathcal{A}^{-}_i$ represent the positive samples in $M$ with the same label as $\mathbf{z}_i$ and the negative samples searched from $M$, respectively. $t$ is a temperature parameter.

\subsubsection{Total loss and algorithm} 
The total training loss of our method can be computed as
\begin{equation}
    \mathcal{L}_{all} =  \mathcal{L}_{ugm}  +\lambda\mathcal{L}_{unk} +\mathcal{L}_{sup}
\end{equation}
where $\lambda$ is a weighting parameter. Moreover, the full algorithm of our method is provided in {\bf Algorithm 1}.


\section{Experimental Results}

\begin{table*}[t]
    \centering
    \resizebox{1\textwidth}{48mm}{
\begin{tabular}{l|cccccccccccc|c|c}
\hline \multicolumn{15}{c}{{\bf Open-partial Domain Adaptation Setting (H-score)}}\\

\hline
\multirow{2}{*}{ Method }& \multicolumn{12}{c|}{ OfficeHome (10/5/50) }& \multicolumn{1}{l|}{}& \multicolumn{1}{c}{VisDA(6/3/3)} \\[0.5pt]
& A2C & A2P & A2R & C2A & C2P & C2R & P2A & P2C & P2R & R2A & R2C & R2P & Avg & S2R \\[0.5pt]
\hline OSBP \cite{saito2018open} & $39.6$ & $45.1$ & $46.2$ & $45.7$ & $45.2$ & $46.8$ & $45.3$ & $40.5$ & $45.8$ & $45.1$ & $41.6$ & $46.9$ & $44.5$ & $27.3$ \\[0.5pt]
UAN \cite{you2019universal} & $51.6$ & $51.7$ & $54.3$ & $61.7$ & $57.6$ & $61.9$ & $50.4$ & $47.6$ & $61.5$ & $62.9$ & $52.6$ & $65.2$ & $56.6$ & $30.5$\\[0.5pt]
CMU \cite{fu2020learning} & $56.0$ & $56.9$ & $59.1$ & $66.9$ & $64.2$ & $67.8$ & $54.7$ & $51.0$ & $66.3$ & $68.2$ & $57.8$ & $69.7$ & $61.6$ & $34.6$\\[0.5pt]
DCC \cite{li2021DCC} & $57.9$ & $54.1$ & $58.0$ & $\mathbf{74.6}$ & $70.6$ & $77.5$ & $64.3$ & $\mathbf{73.6}$ & $74.9$ & $\mathbf{80.9}$ & $\mathbf{75.1}$ & $80.4$ & $70.1$ & $43.0$\\[0.5pt]
OVANet \cite{saito2021ovanet} & ${6 2 . 8}$ & $75.6$ & $78.6$ & $70.7$ & $68.8$ & $75.0$ & $71.3$ & $58.6$ & $80.5$ & $76.1$ & $64.1$ & $78.9$ & $71.8$& $53.1$ \\[0.5pt]
GATE \cite{chen2022geometric} & ${6 3 . 8}$ & $75.9$ & $81.4$ & $74.0$ & $\mathbf{72.1}$ & $\mathbf{79.8}$ & $\mathbf{74.7}$ & $70.3$ & $82.7$ & $79.1$ & $71.5$ & $81.7$ & $75.6$& $\mathbf{56.4}$ \\[0.5pt]
\hline 
\rowcolor{gray!30}  Ours & $\mathbf{6 4 . 5}$ & $\mathbf{77.9}$ & $\mathbf{87.1}$ & $74.1$ & ${71.0}$ & ${78.1}$ & ${7 4.2}$ & $70.0$ & $\mathbf{8 4 . 1}$ & $80 . 2$ & $71.5$ & $\mathbf{86.2}$ & $\mathbf{7 6 . 6}$& $55.3$ \\
\hline \multicolumn{15}{c}{{\bf Open-set Domain Adaptation Setting (H-score)}}\\
\hline \multirow{2}{*}{ Method } & \multicolumn{12}{c|}{ OfficeHome (25/0/40) } & \multicolumn{1}{l|}{}& \multicolumn{1}{c}{VisDA(6/0/6)} \\[0.5pt]
& A2C & A2P & A2R & C2A & C2P & C2R & P2A & P2C & P2R & R2A & R2C & R2P & Avg & S2R  \\[0.5pt]
\hline
OSBP \cite{saito2018maximum} & $55.1$ & $65.2$ & $72.9$ & $64.3$ & $64.7$ & $70.6$ & $63.2$ & $53.2$ & $73.9$ & $66.7$ & $54.5$ & $72.3$ & $64.7$ & $52.3$ \\[0.5pt]
ROS \cite{bucci2020effectiveness} & $60.1$ & $69.3$ & $76.5$ & $58.9$ & $65.2$ & $68.6$ & $60.6$ & $56.3$ & $74.4$ & $68.8$ & $60.4$ & $75.7$ & $66.2$ & $66.5$ \\[0.5pt]
UAN \cite{you2019universal} & $40.3$ & $41.5$ & $46.1$ & $53.2$ & $48.0$ & $53.7$ & $40.6$ & $39.8$ & $52.5$ & $53.6$ & $43.7$ & $56.9$ & $47.5$ & $51.9$ \\[0.5pt]
CMU \cite{fu2020learning} & $61.9$ & $61.3$ & $63.7$ & $64.2$ & $58.6$ & $62.6$ & $67.4$ & $61.0$ & $65.5$ & $65.9$ & $61.3$ & $64.2$ & $63.0$ & $67.5$ \\[0.5pt]
DCC \cite{li2021DCC} & $56.1$ & $67.5$ & $66.7$ & $49.6$ & $66.5$ & $64.0$ & $55.8$ & $53.0$ & $70.5$ & $61.6$ & $57.2$ & $71.9$ & $61.7$ & $59.6$\\[0.5pt]
OVANet \cite{saito2021ovanet} & $58.9$ & $66.0$ & $70.4$ & $62.2$ & $65.7$ & $67.8$ & $60.0$ & $52.6$ & $69.7$ & $68.2$ & $59.1$ & $67.6$ & $64.0$ & $66.1$\\[0.5pt]
GATE \cite{chen2022geometric} & $63.8$ & $70.5$ & $75.8$ & $66.4$ & $67.9$ & $71.7$ & $67.3$ & $61.5$ & $76.0$ & $70.4$ & $61.8$ & $75.1$ & $69.1$ & $70.8$\\[0.5pt]
\hline 
\rowcolor{gray!30}  Ours & $6 2 . 5$ & $\mathbf{79.1}$ & $\mathbf{80.9}$ & $72.2$ & $\mathbf{71.7}$ & $\mathbf{78.5}$ & $\mathbf{7 3.6}$ & $61.8$ & $\mathbf{8 4 . 5}$ & $7 9 . 1$ & $6 5.7$ & $\mathbf{82.8}$ & ${\mathbf{7 4 . 4}}$ & $\mathbf{74.5}$\\
\hline
\end{tabular}}
    \caption{Comparison of main results on OfficeHome. Some results for previous methods are cited from OVANet \cite{saito2021ovanet} and {GATE \cite{chen2022geometric}.}}
    \label{tab:officehomeopda}
\end{table*}

\begin{figure*}[t]
\centering

\subfigure{
\begin{minipage}[t]{0.2\linewidth}
\centering
\includegraphics[width=1.44in]{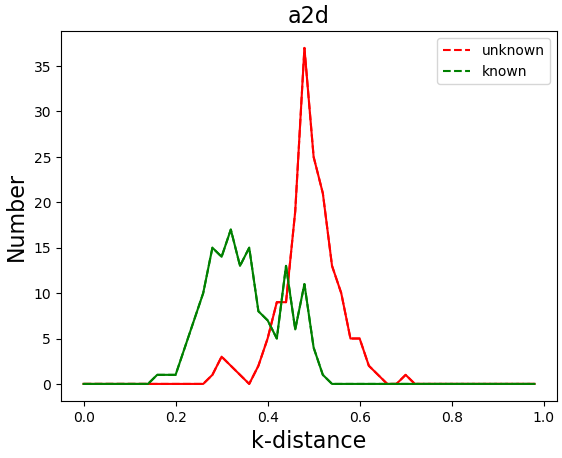}
\end{minipage}%

\begin{minipage}[t]{0.2\linewidth}
\centering
\includegraphics[width=1.44in]{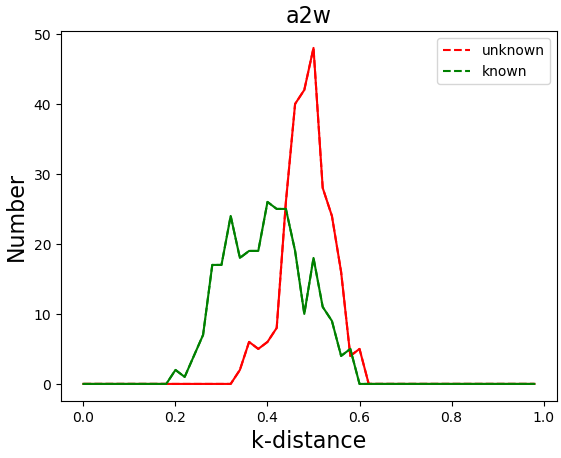}
\end{minipage}%
\begin{minipage}[t]{0.2\linewidth}
\centering
\includegraphics[width=1.44in]{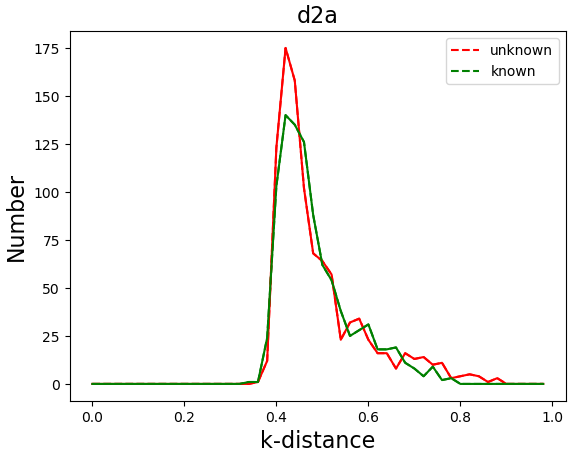}
\end{minipage}%
\begin{minipage}[t]{0.2\linewidth}
\centering
\includegraphics[width=1.44in]{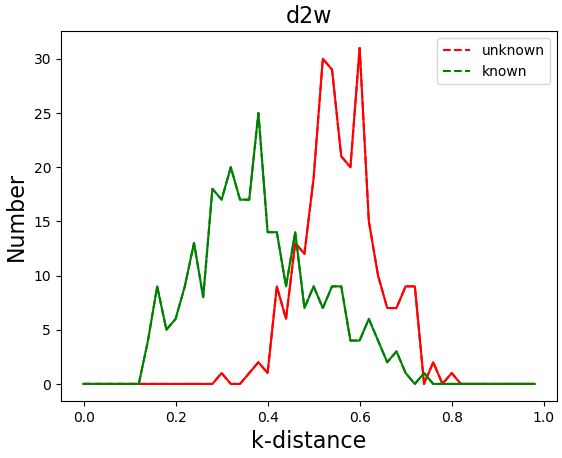}
\end{minipage}%
\begin{minipage}[t]{0.2\linewidth}
\centering
\includegraphics[width=1.44in]{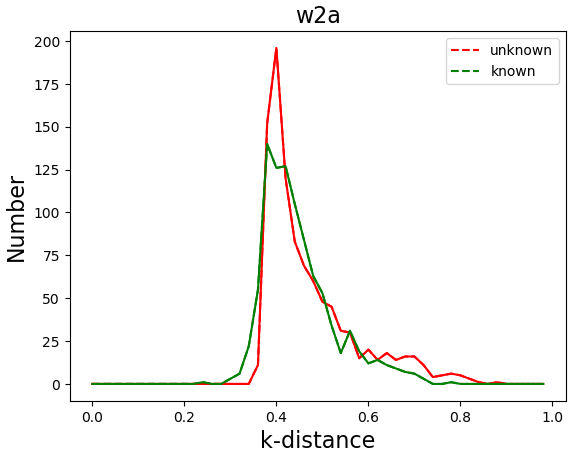}
\end{minipage}%
}%
\\
\subfigure{
\begin{minipage}[t]{0.2\linewidth}
\centering
\includegraphics[width=1.44in]{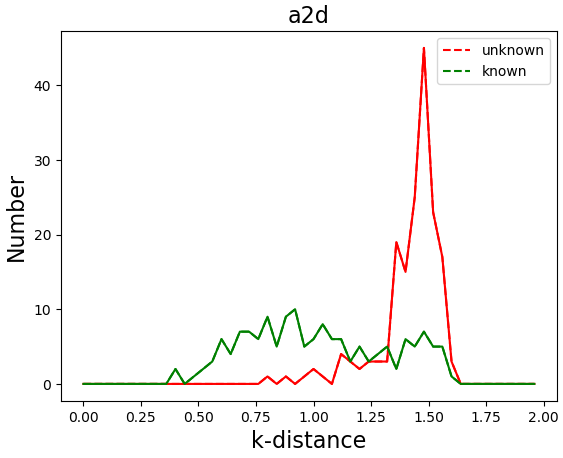}
\end{minipage}%
\begin{minipage}[t]{0.2\linewidth}
\centering
\includegraphics[width=1.44in]{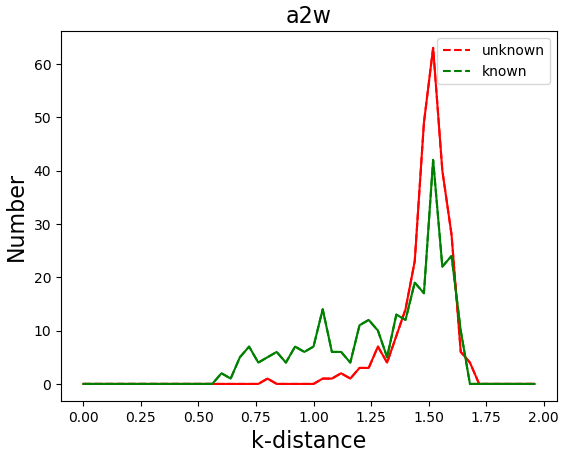}
\end{minipage}%
\begin{minipage}[t]{0.2\linewidth}
\centering
\includegraphics[width=1.44in]{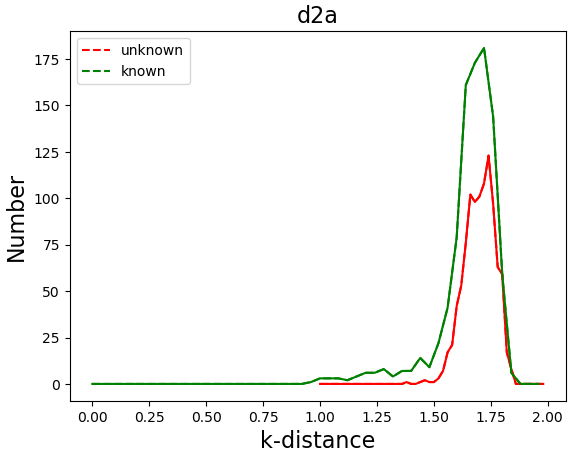}
\end{minipage}%
\begin{minipage}[t]{0.2\linewidth}
\centering
\includegraphics[width=1.44in]{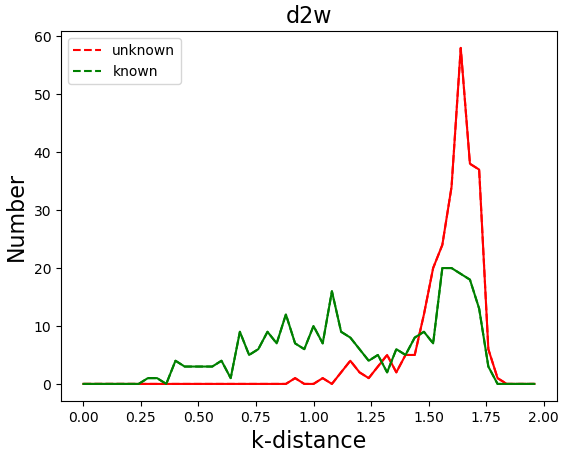}
\end{minipage}%
\begin{minipage}[t]{0.2\linewidth}
\centering
\includegraphics[width=1.44in]{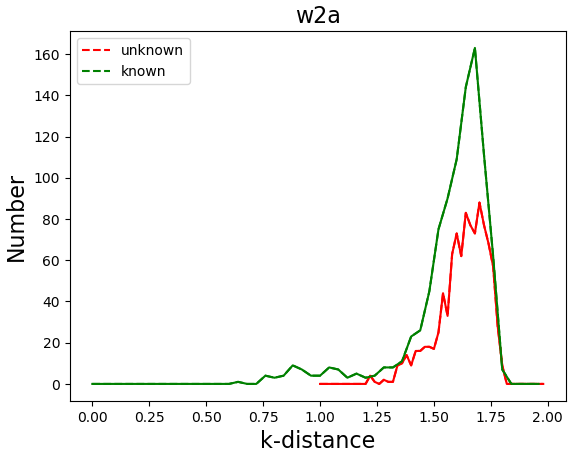}
\end{minipage}%
}%
\vspace{-0.45cm}
\centering
\caption{Graphs of distributions of $k$-nearest neighbor distances of target samples in an early epoch on Office-$31$ with the OPDA setting. The first row represents the distribution in the original feature space and the second row represents that in the subspace.  The \textcolor{red}{red} lines represent the distribution of unknown target samples while the \textcolor{green}{green} lines represent that of known target samples.}
\label{kdis}
\end{figure*}

\begin{figure*}[t]
\centering

\subfigure{
\begin{minipage}[t]{0.2\linewidth}
\centering
\includegraphics[width=1.44in]{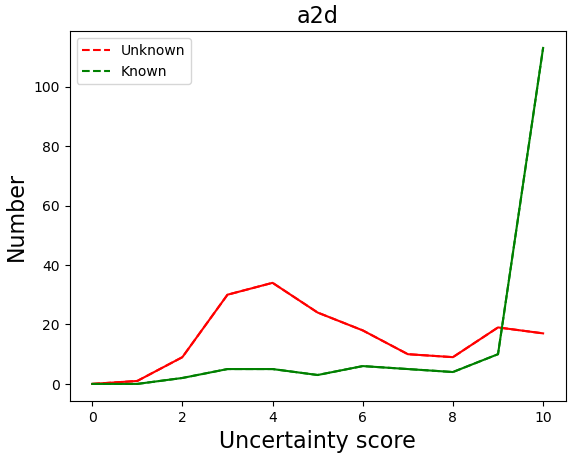}
\end{minipage}%
\begin{minipage}[t]{0.2\linewidth}
\centering
\includegraphics[width=1.44in]{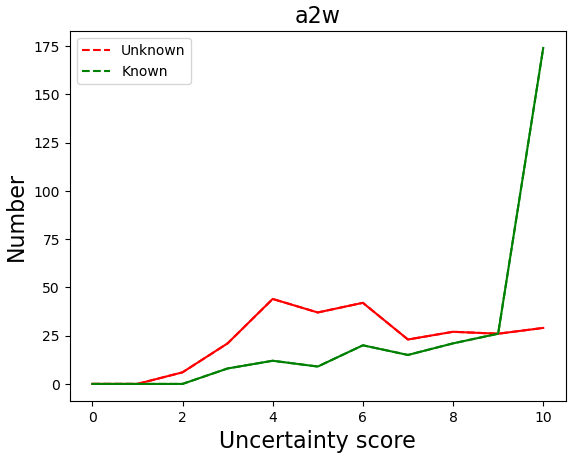}
\end{minipage}%
\begin{minipage}[t]{0.2\linewidth}
\centering
\includegraphics[width=1.44in]{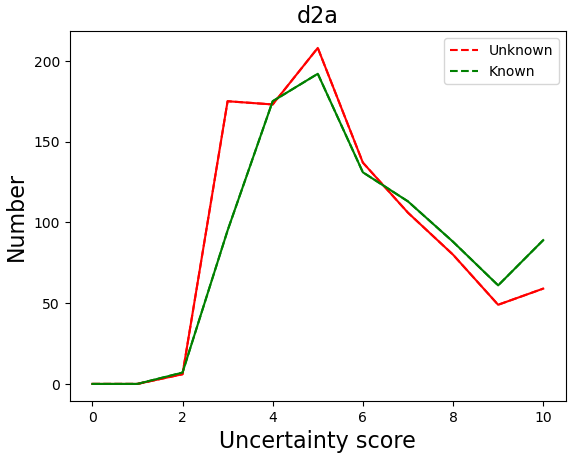}
\end{minipage}%
\begin{minipage}[t]{0.2\linewidth}
\centering
\includegraphics[width=1.44in]{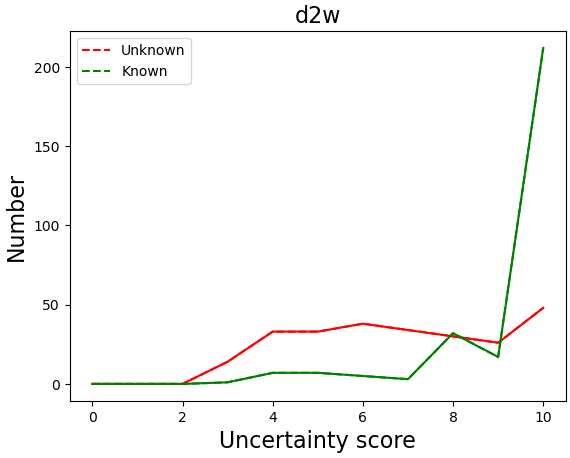}
\end{minipage}%
\begin{minipage}[t]{0.2\linewidth}
\centering
\includegraphics[width=1.44in]{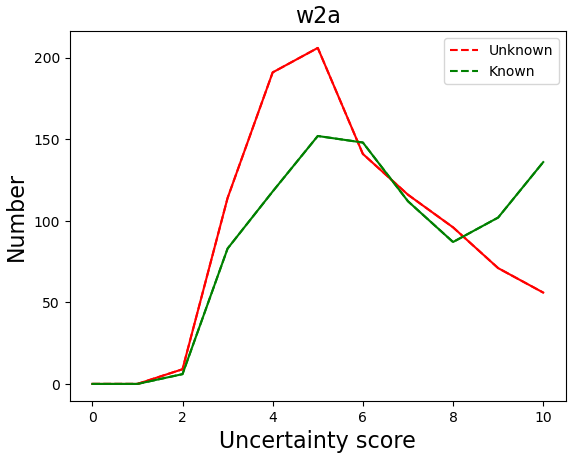}
\end{minipage}%
}%
\hspace{-0.1cm}
\subfigure{
\begin{minipage}[t]{0.2\linewidth}
\centering
\includegraphics[width=1.44in]{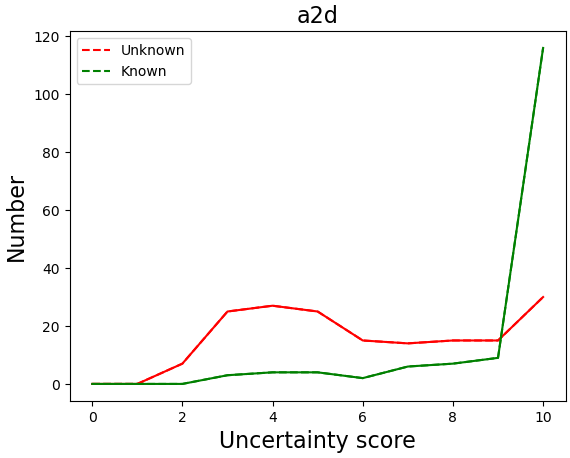}
\end{minipage}%
\begin{minipage}[t]{0.2\linewidth}
\centering
\includegraphics[width=1.44in]{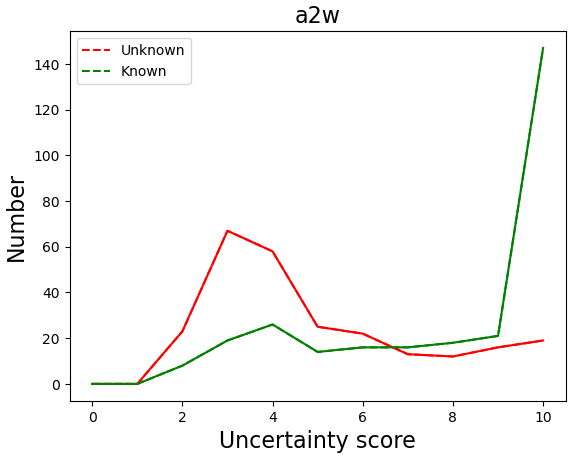}
\end{minipage}%
\begin{minipage}[t]{0.2\linewidth}
\centering
\includegraphics[width=1.44in]{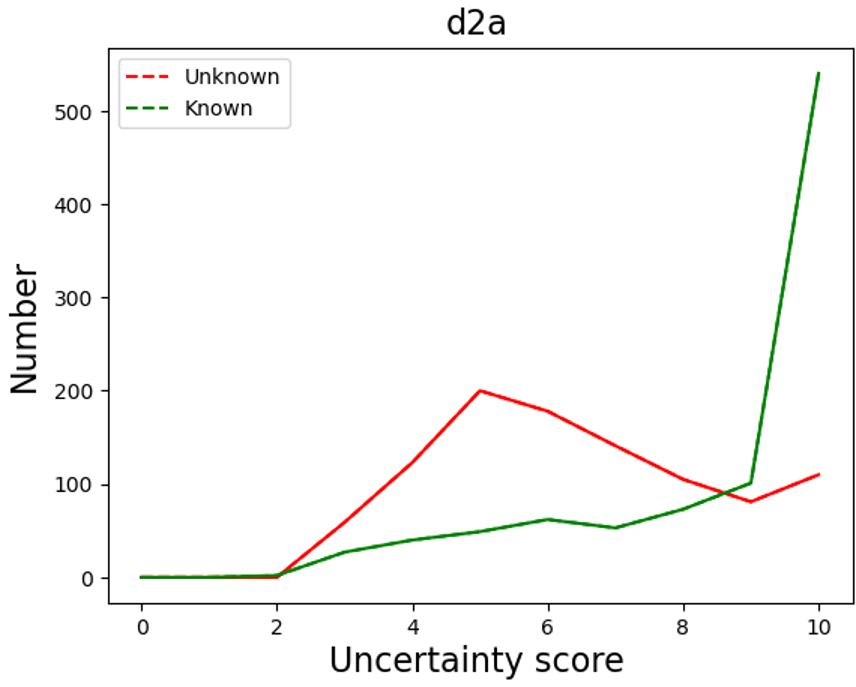}
\end{minipage}%
\begin{minipage}[t]{0.2\linewidth}
\centering
\includegraphics[width=1.44in]{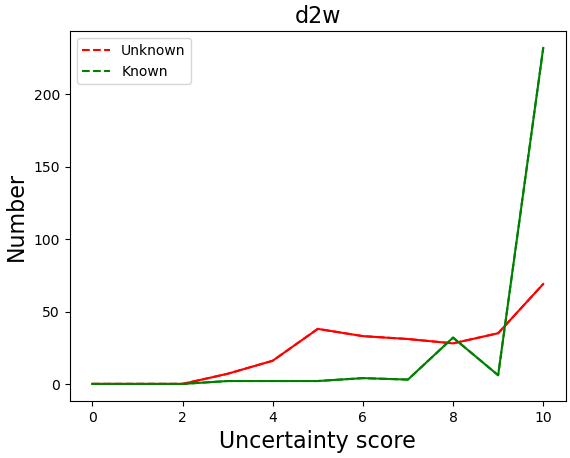}
\end{minipage}%
\begin{minipage}[t]{0.2\linewidth}
\centering
\includegraphics[width=1.44in]{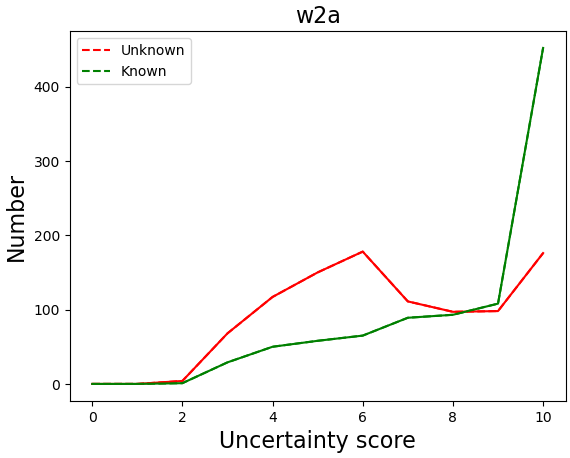}
\end{minipage}%
}%
\vspace{-0.45cm}
\centering
\caption{ Graphs of distributions of the maximum numbers of neighbors belonging to the same class of target samples in an early epoch on Office-$31$ with the OPDA setting. The first row represents the distribution in the original feature space and the second row represents that in the subspace. The \textcolor{red}{red} lines represent the distribution of unknown target samples while the  \textcolor{green}{green} lines represent that of known target samples.}
\label{con}
\end{figure*}

\begin{figure*}[t]
\centering
\vspace{-0.3cm}
\subfigure{
\begin{minipage}[t]{0.2\linewidth}
\centering
\includegraphics[width=1.44in]{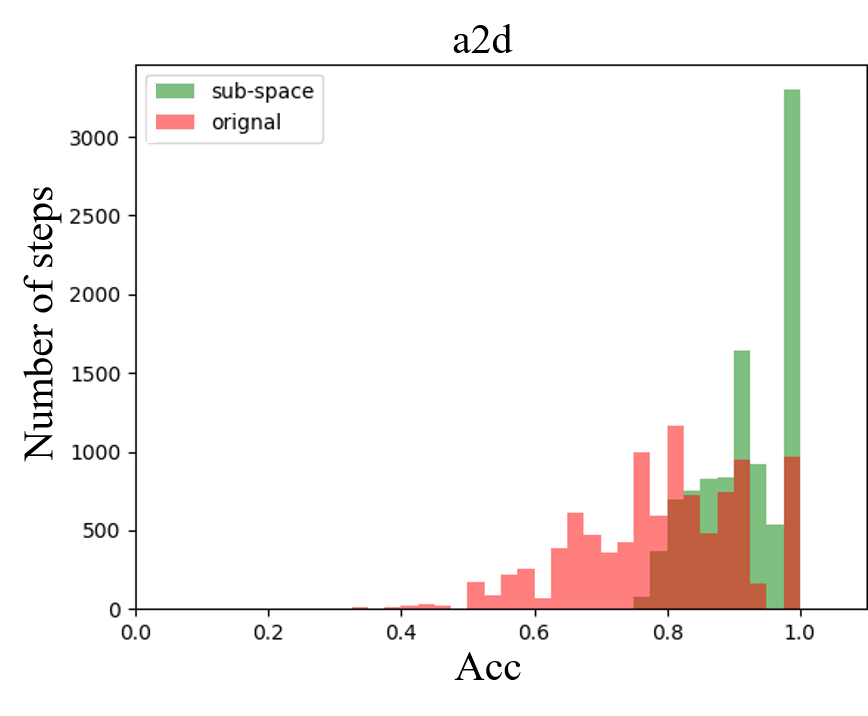}
\end{minipage}%
\begin{minipage}[t]{0.2\linewidth}
\centering
\includegraphics[width=1.44in]{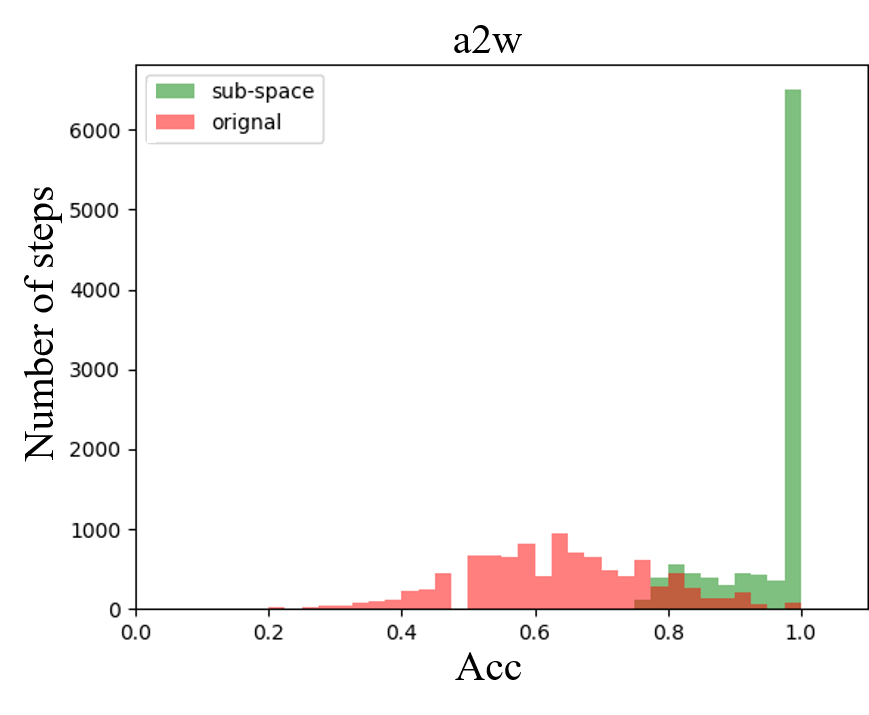}
\end{minipage}%
\begin{minipage}[t]{0.2\linewidth}
\centering
\includegraphics[width=1.44in]{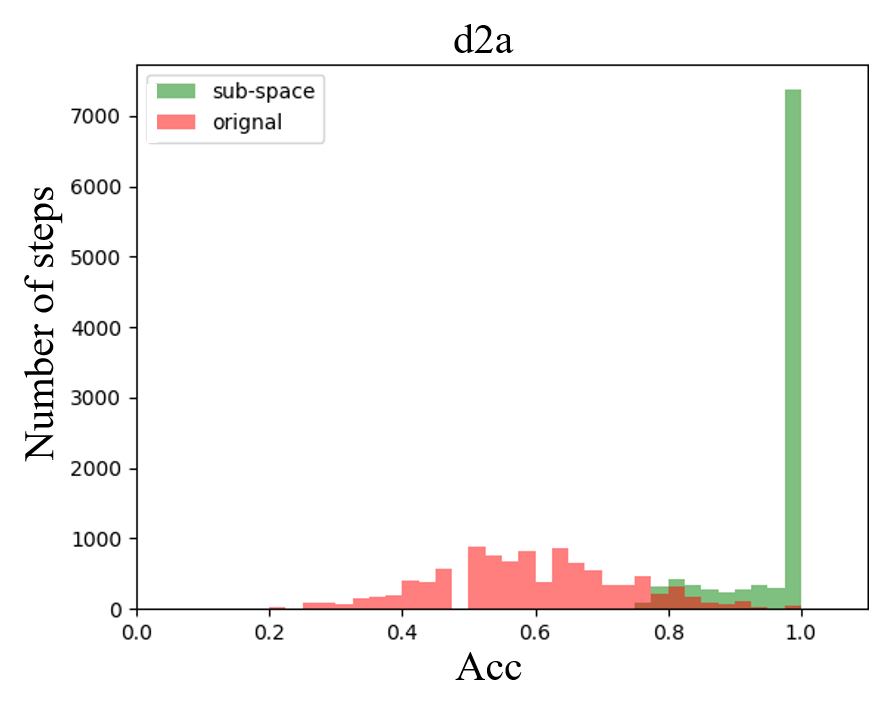}
\end{minipage}%
\begin{minipage}[t]{0.2\linewidth}
\centering
\includegraphics[width=1.44in]{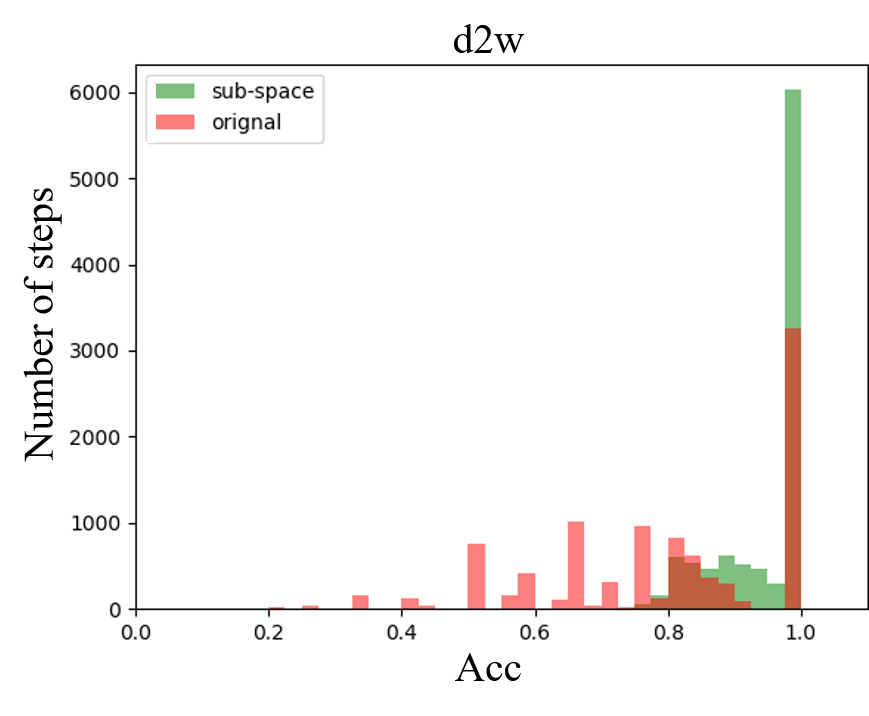}
\end{minipage}%
\begin{minipage}[t]{0.2\linewidth}
\centering
\includegraphics[width=1.44in]{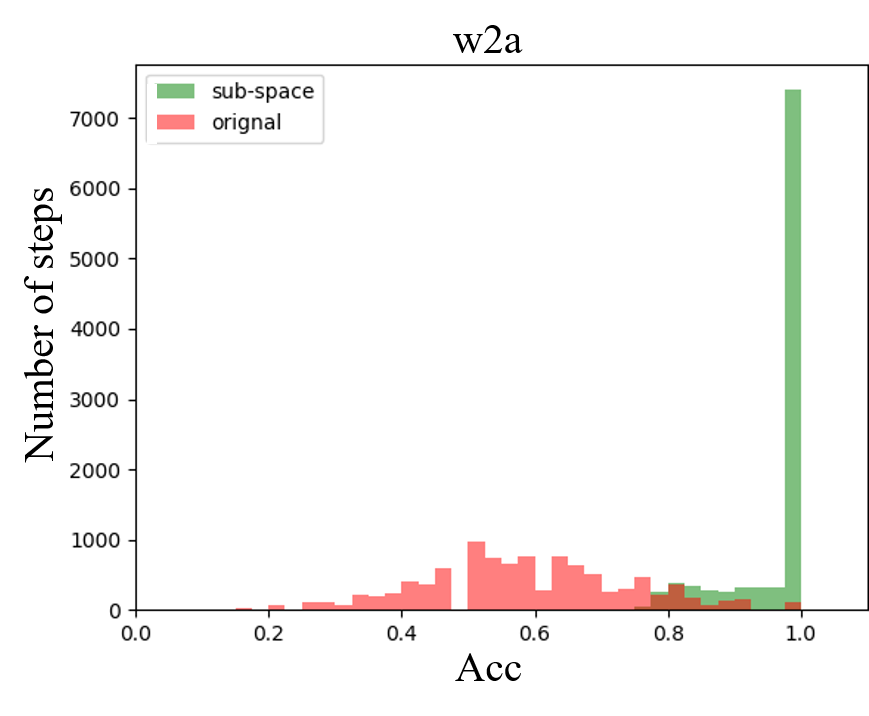}
\end{minipage}%
}%
\hspace{-0.1cm}
\subfigure{
\begin{minipage}[t]{0.2\linewidth}
\centering
\includegraphics[width=1.44in]{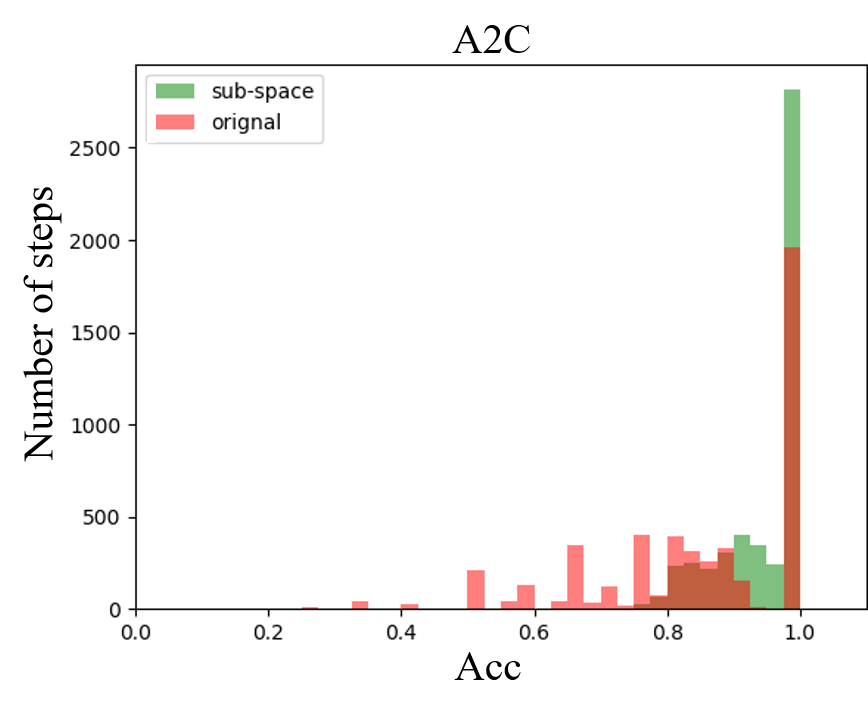}
\end{minipage}%
\begin{minipage}[t]{0.2\linewidth}
\centering
\includegraphics[width=1.44in]{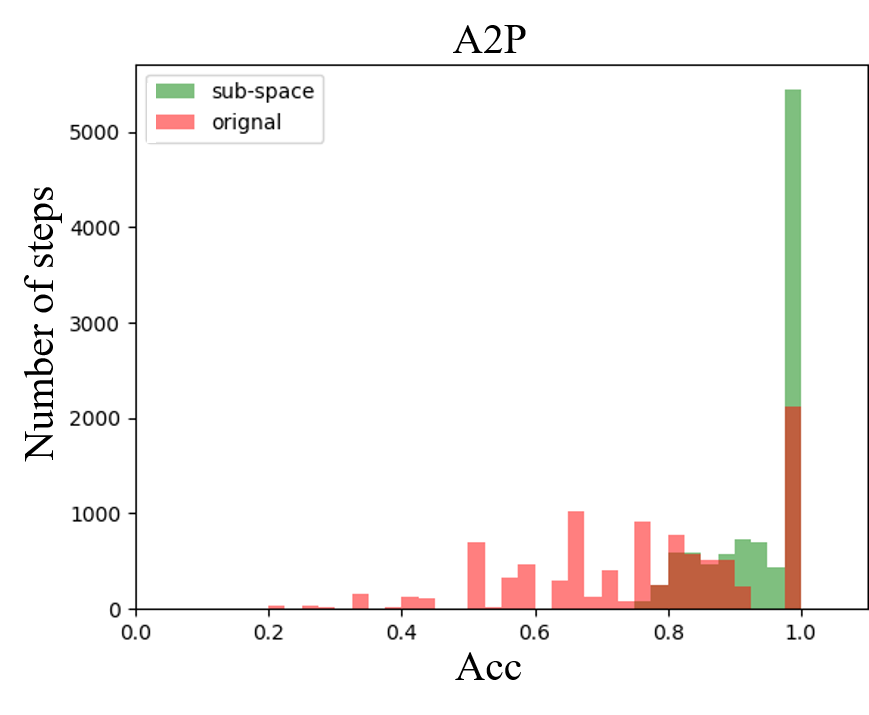}
\end{minipage}%
\begin{minipage}[t]{0.2\linewidth}
\centering
\includegraphics[width=1.44in]{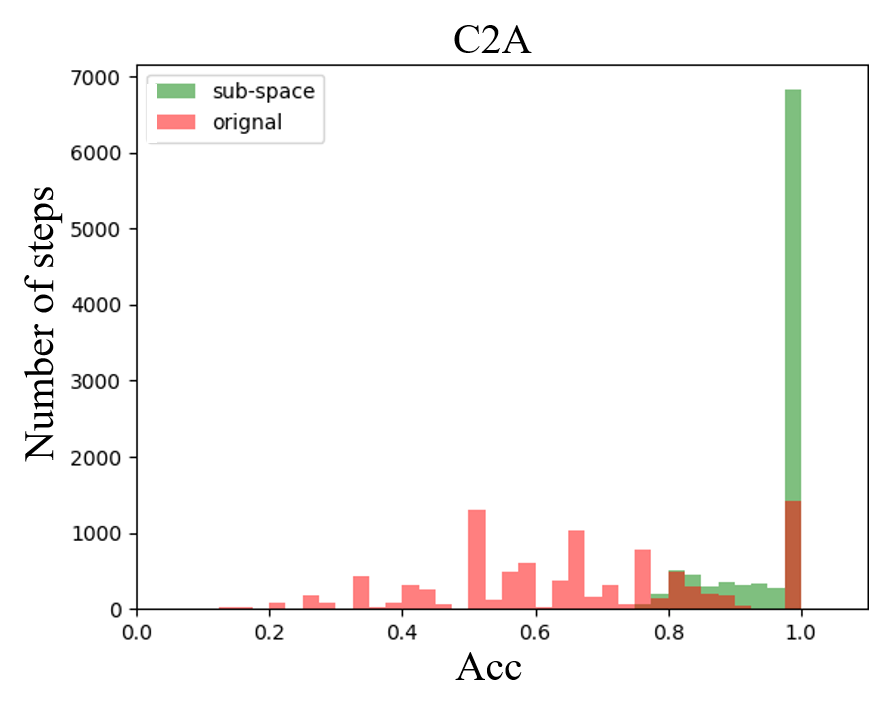}
\end{minipage}%
\begin{minipage}[t]{0.2\linewidth}
\centering
\includegraphics[width=1.44in]{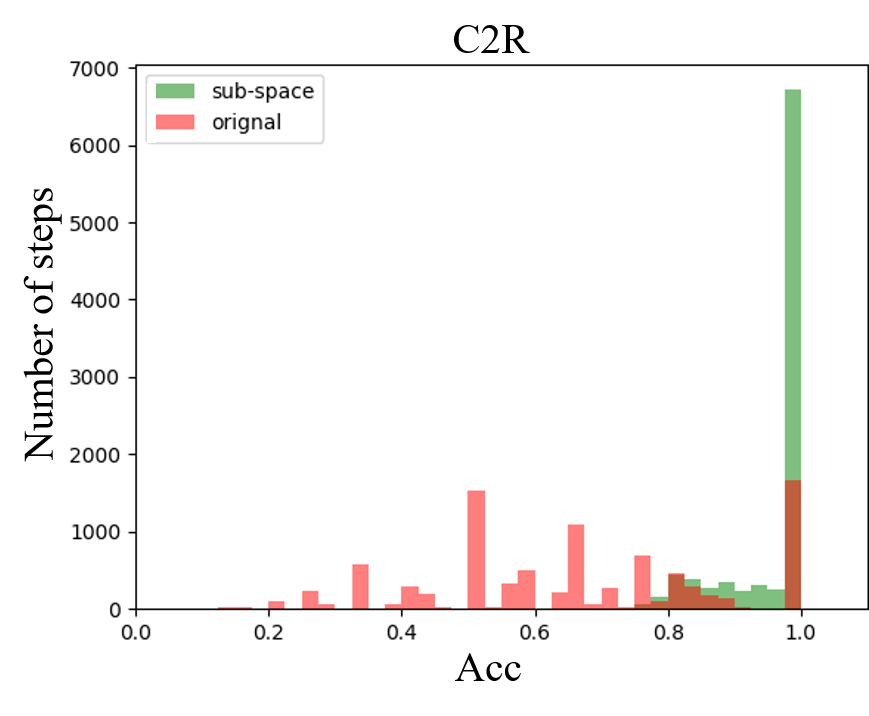}
\end{minipage}%
\begin{minipage}[t]{0.2\linewidth}
\centering
\includegraphics[width=1.44in]{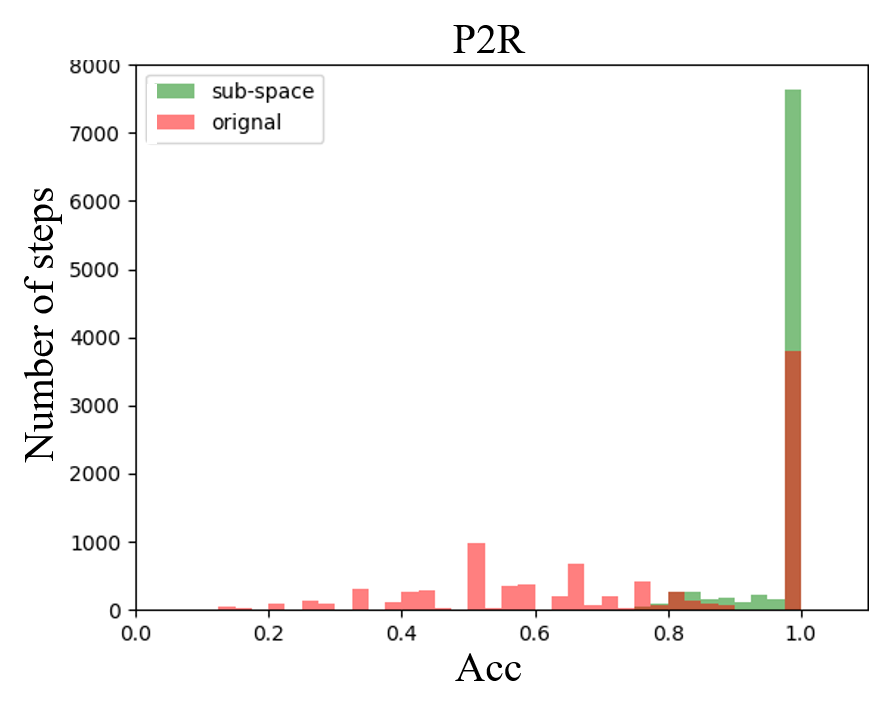}
\end{minipage}%
}%
\vspace{-0.35cm}
\centering
\caption{ 
Histograms of accuracy of  {unknown samples discovering} in the original feature space (\textcolor{red}{red} columns) and in the subspace (\textcolor{green}{green} columns) on Office-$31$ (a$2$d, a$2$w, d$2$a, d$2$w and w$2$a) and OfficeHome (A$2$C, A$2$P, C$2$A C$2$R and P$2$R) with the OPDA setting.}
\label{acc}
\end{figure*}

In this section, we first introduce our experimental setups including datasets, evaluation protocols and training details. Then, we introduce some baselines of recent SOTA methods in UniDA and compare results in the main datasets (i.e. Office-31, OfficeHome and VisDA) with them. We also conduct extensive ablation studies to demonstrate the effectiveness of each component of the proposed method. All experiments were implemented on one RTX2080Ti 11GB GPU with PyTorch 1.7.1 \cite{paszke2019pytorch}. 

\subsection{Experimental Setups}
\subsubsection{Datasets and evaluation protocols}
We conduct experiments on four datasets. {\bf Office-31} \cite{saenko2010adapting} consists of $4,652$ images from three domains: DSLR (D), Amazon (A), and Webcam (W). {\bf OfficeHome} \cite{peng2019moment} is a more  challenging dataset, which consists of $15,500$ images from $65$ categories. It is made up of $4$ domains: Artistic images (A), Clip-Art images (C), Product images (P), and Real-World images (R). {\bf VisDA} \cite{peng2017visda} is a large-scale dataset, where the source domain contains $15,000$ synthetic images and the target domain consists of $5,000$ images from the real world.

In this paper, we use the \textbf{H-score} in line with recent UniDA methods \cite{fu2020learning,li2021DCC,saito2021ovanet}. H-score, proposed by Fu \textit{et al.} \cite{fu2020learning}, is the harmonic mean of the accuracy on the common classes $a_{com}$ and the accuracy on the unknown class $a_{unk}$:
\begin{equation}
    h = \frac{2a_{com}  a_{unk}}{a_{com} + a_{unk}}.
    \label{h-score}
\end{equation}

\subsubsection{Training details} We employ the ResNet-50 \cite{he2016deep} backbone pre-trained on ImageNet\cite{deng2009imagenet}, and optimize the model using Nesterov momentum SGD with momentum of $0.9$ and weight decay of $5 \times 10^{-4}$ . The batch size is set to $36$ through all datasets for both domains. The initial learning rate is set as $0.01$ for the classifier layers and $0.001$ for the backbone layers. The learning rate is decayed with the inverse learning rate decay scheduling.  The number of neighbors retrieved is set to be dependent on the sizes of the datasets. For Office-31 ($4,652$ images in $31$ categories) and OfficeHome ($15,500$ images in $65$ categories), the number of retrieved neighbors $|\mathcal{N}_i|$ is set to $10$. For VisDA ($20,000$ images in total), we set $|\mathcal{N}_i|$ to $100$ , respectively. We set $\lambda$ to  $0.1$ and $t$ to $0.05$  which are constant through all the datasets. In the test phase, the threshold of distinguishing the unknown samples is set to $\frac{\log C}{2}$ following DANCE~\cite{saito2020universal} which used the entropy of the classifier's output to determine the unknown samples.

\subsection{Comparison With the SOTA Methods}
\subsubsection{Baselines}  We aim to show that our method can better balance the confidences of known and unknown samples for UniDA by comparing our method with the current SOTA methods, such as UAN~\cite{you2019universal} and DANCE~\cite{saito2020universal}, which employed a softmax-based classifier to produce the confidence of each sample to determine whether it belongs to the unknown class or not. Also, we compare our method with DANCE~\cite{saito2020universal}, DCC~\cite{saito2021ovanet}  and GATE~\cite{chen2022geometric} to show that it is better able to solve the domain misalignment by projecting features into a linear subspace  rather than operating in the original feature space.

\begin{figure*}[t]

\subfigure[]{
\begin{minipage}[t]{0.32\linewidth}
\centering
\includegraphics[width=2.2in]{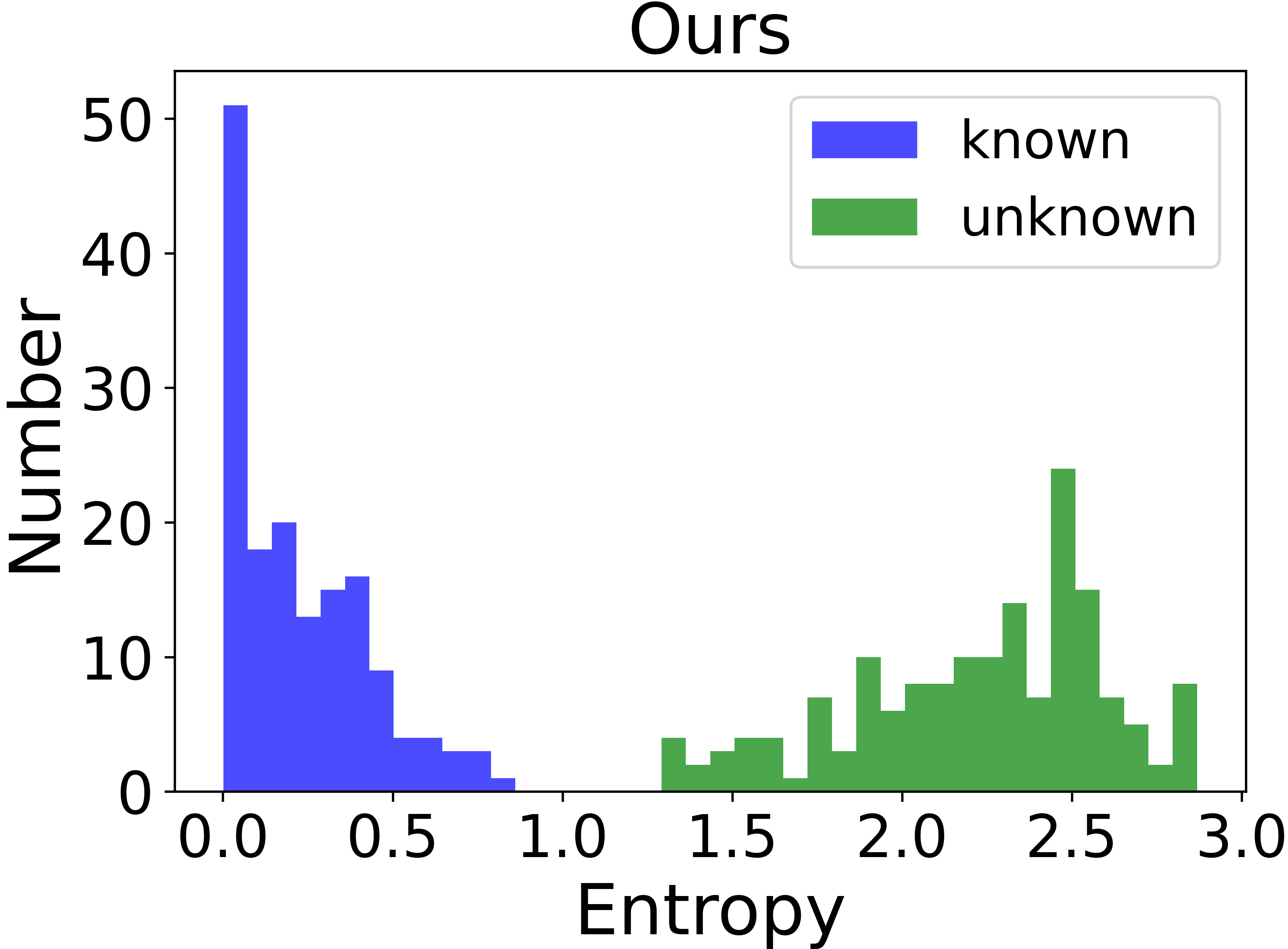}
\end{minipage}%
}%
\hspace{-1mm}
\subfigure[]{
\begin{minipage}[t]{0.32\linewidth}
\centering
\includegraphics[width=2.2in]{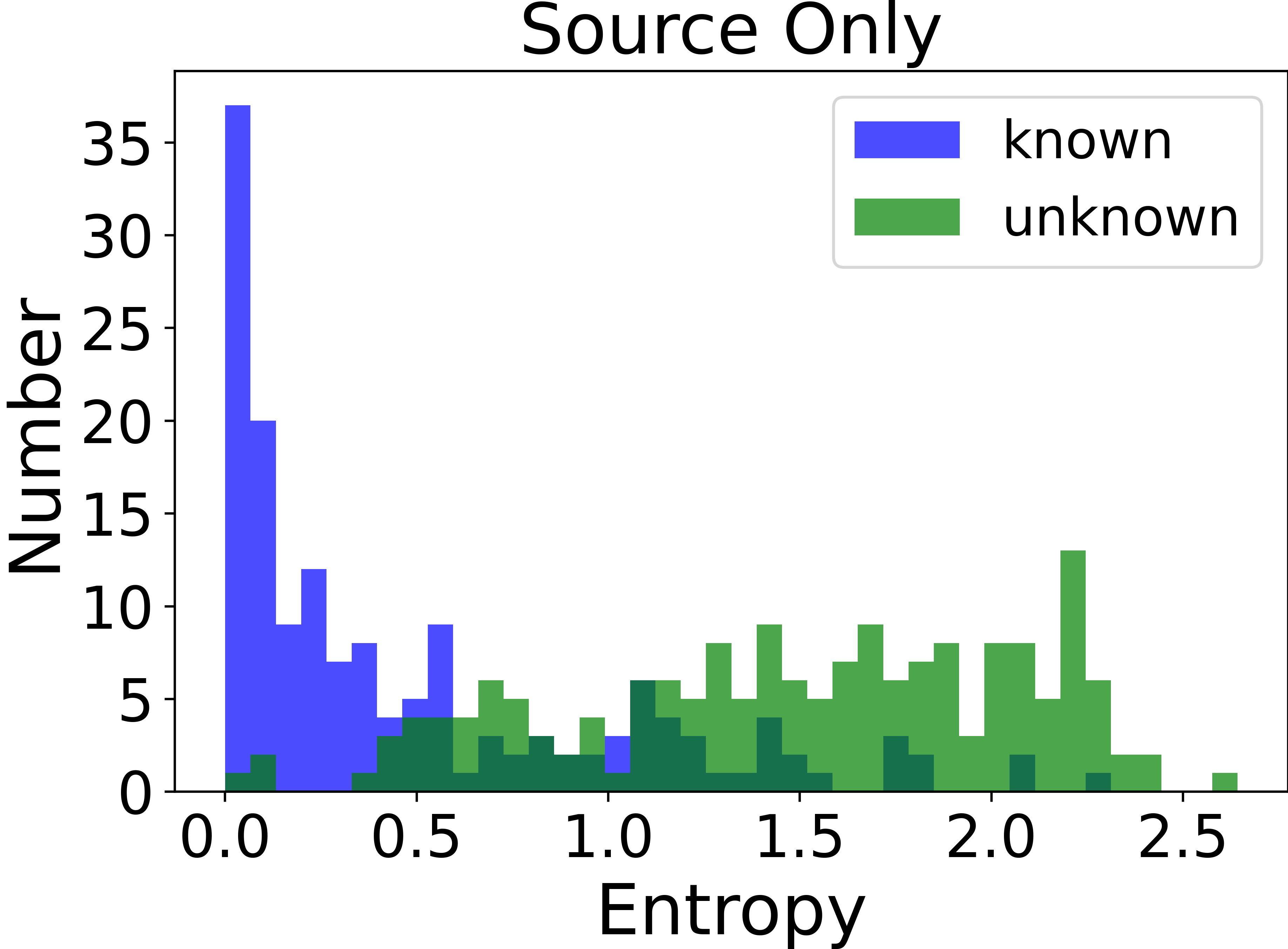}
\end{minipage}%
}%
\hspace{-1mm}
\subfigure[]{
\begin{minipage}[t]{0.32\linewidth}
\centering
\includegraphics[width=2.2in]{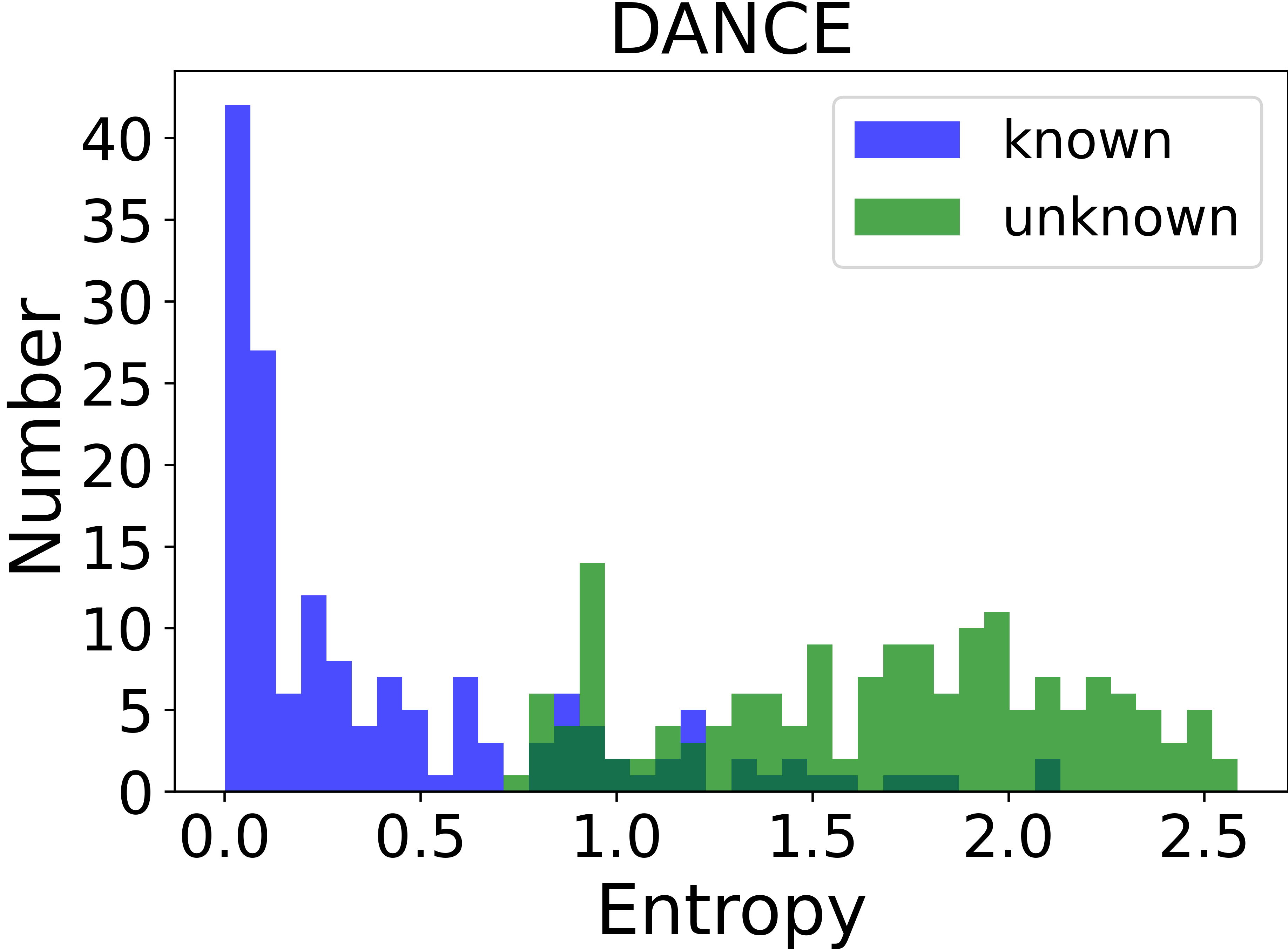}
\end{minipage}%
}%

\centering
\caption{{Comparison on the distribution of the entropy. The three plots of histograms show the entropy at the last epoch produced by the full version of our method, the model trained only on source domain, and the model trained on DANCE~\cite{saito2020universal} in Office-31(A2D) respectively. Each area in dark green indicates that there is an overlap between the \textcolor{green}{green} and the \textcolor{blue}{blue} bars}.}
\label{ed}
\end{figure*}

\subsubsection{Results in main datasets} TABLE \ref{officeopda} lists the results on Office-31 with the OPDA and ODA settings respectively. TABLE \ref{tab:officehomeopda} lists the results on OfficeHome and VisDA both with the OPDA and ODA settings, respectively.  On Office-31, our method outperforms the SOTA methods by $1.2\%$ in terms of the H-score on average with the OPDA setting,  and makes a significant improvement of $2.0\%$ in terms of the H-score on average with the ODA setting. For the more challenging dataset OfficeHome which contains much more private classes than common classes, our method also makes a significant improvement of $5.3\%$ in terms of the H-score with the ODA setting.
Our method consistently performs better than other methods. VisDA is a much larger dataset than Office-31 and OfficeHome which contains about $10,000$ images in each domain.
Our method achieves the SOTA performance  on VisDA with the significant improvement of $3.7\%$ in the ODA setting.  

\subsubsection{Summary} 
According to the results of the quantitative comparisons, our method achieves the SOTA performance in every dataset and most subtasks, which demonstrates the effectiveness of
the main idea of our method that solves the domain misalignment through mapping features into a linear subspace and balances the confidences of known and unknown samples by controlling the intra-class variance of the source domain.

\subsection{Ablation Studies}
\label{ablation}
We provide further analysis and ablations to understand the behavior of each major component of our method in this section.

\subsubsection{A closer look at the  unknown samples discovering scheme}
By Corollary~\ref{corollary}, the posterior probability of a target sample belonging to the unknown class depends on the largest number of neighbors belonging to the same class. Thus, we compare the proposed  {unknown samples discovering} scheme with the discovering method based on the $k$-nearest neighbor distance in this part. Firstly, we collect the $k$-nearest neighbor distances of all target samples in an early epoch in Office-$31$ which can be more influential for the whole training. As shown in the first row in Fig.~\ref{kdis}, the distributions of known samples are not distinguishable enough, especially in a$2$w, d$2$a and w$2$a. It is obvious that the $k$-nearest neighbor distance is not reliable enough to discover the unknown samples. Moreover, the optimal thresholds for each subtask is different and hard to choose. Notably, mapping samples into the linear subspace has a significant influence in making  the distribution of data points  more uniform as shown in the second row in Fig.~\ref{kdis}. However, it is unable to improve the discrimination of unknown samples based on the $k$-nearest neighbor distance. By contrast, the distributions of confidences defined by the largest number of neighbors belonging to the same class illustrated in the first row of Fig.~\ref{con} are much more distinguishable.  



\begin{figure*}[t]
\centering
\subfigure{
\begin{minipage}[t]{0.2\linewidth}
\centering
\includegraphics[width=1.45in]{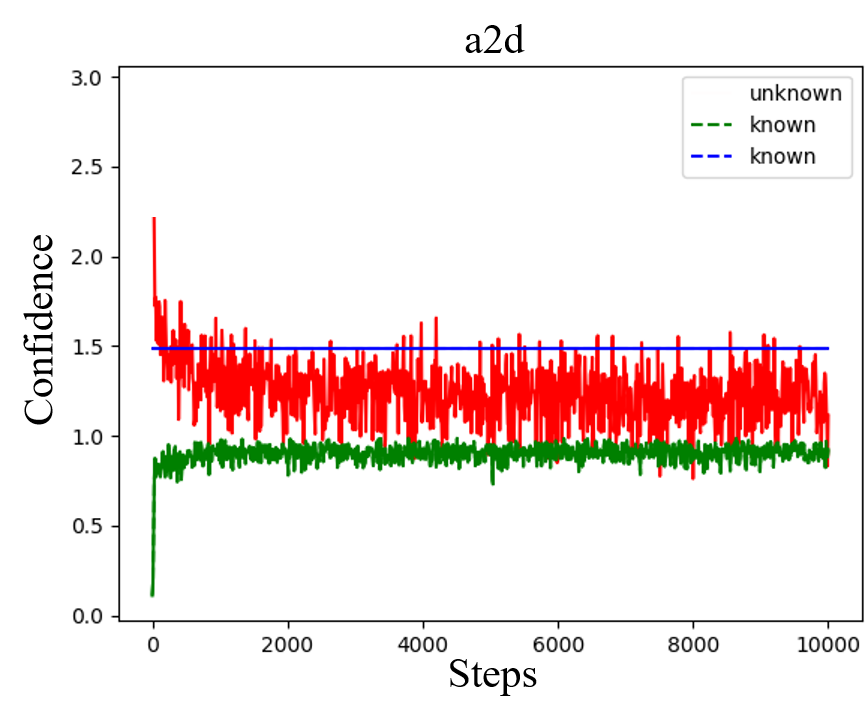}
\end{minipage}%
\begin{minipage}[t]{0.2\linewidth}
\centering
\includegraphics[width=1.45in]{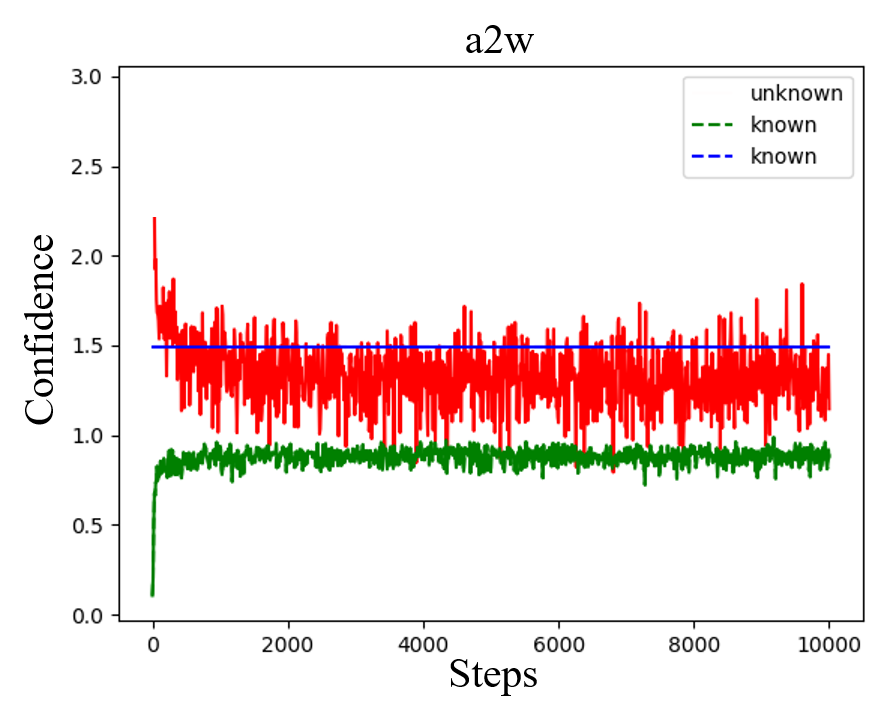}
\end{minipage}%
\begin{minipage}[t]{0.2\linewidth}
\centering
\includegraphics[width=1.45in]{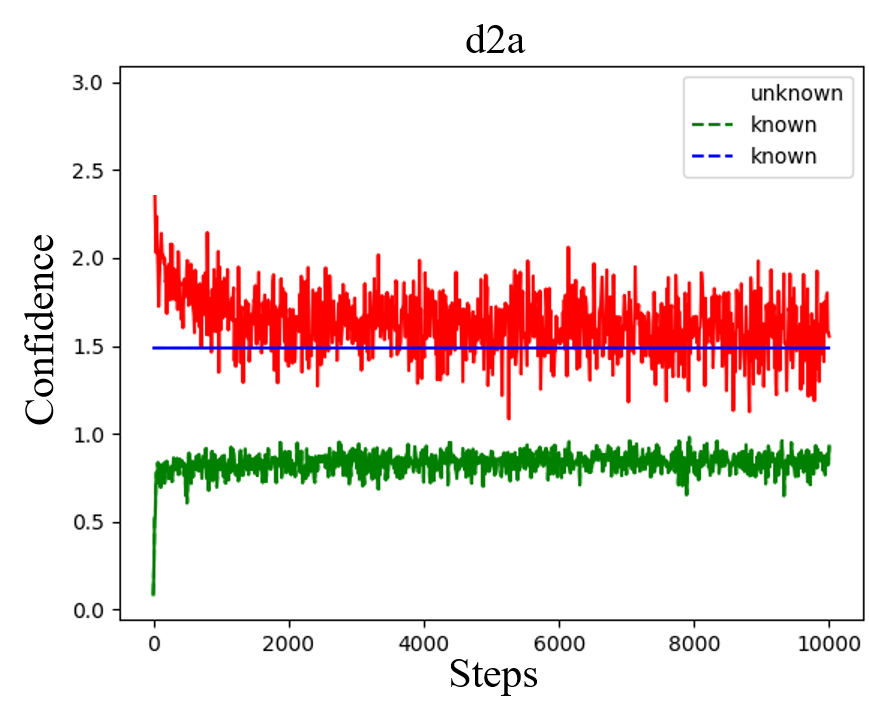}
\end{minipage}%
\begin{minipage}[t]{0.2\linewidth}
\centering
\includegraphics[width=1.45in]{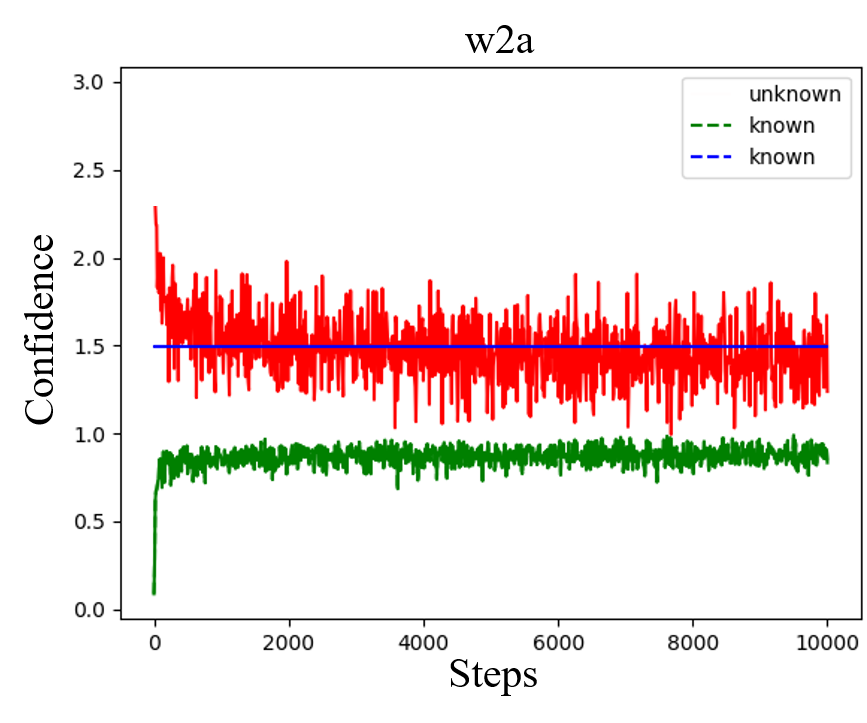}
\end{minipage}%
\begin{minipage}[t]{0.2\linewidth}
\centering
\includegraphics[width=1.45in]{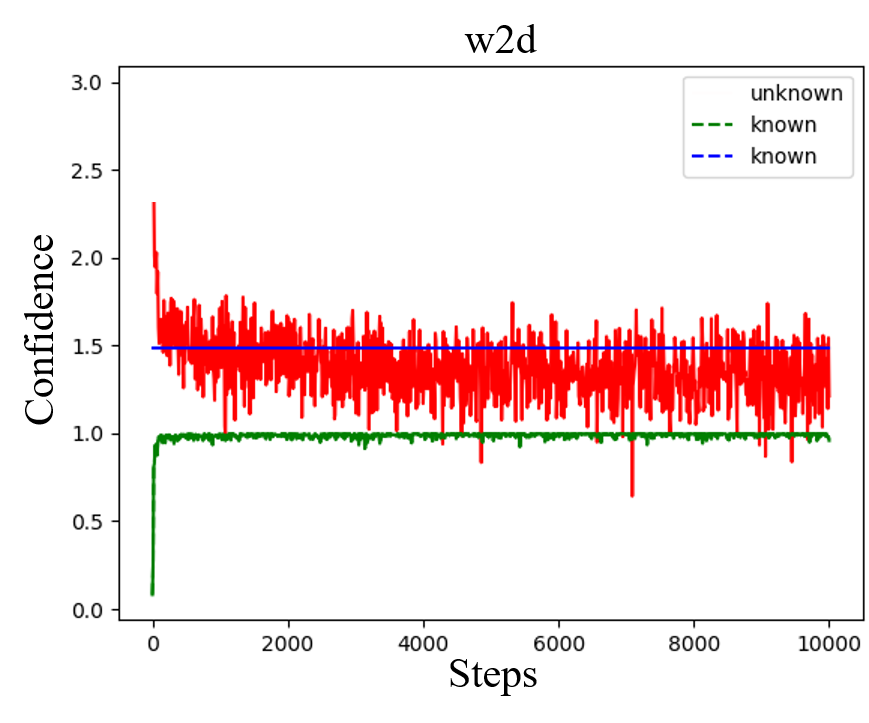}
\end{minipage}%
}%
\\
\subfigure{
\begin{minipage}[t]{0.2\linewidth}
\centering
\includegraphics[width=1.45in]{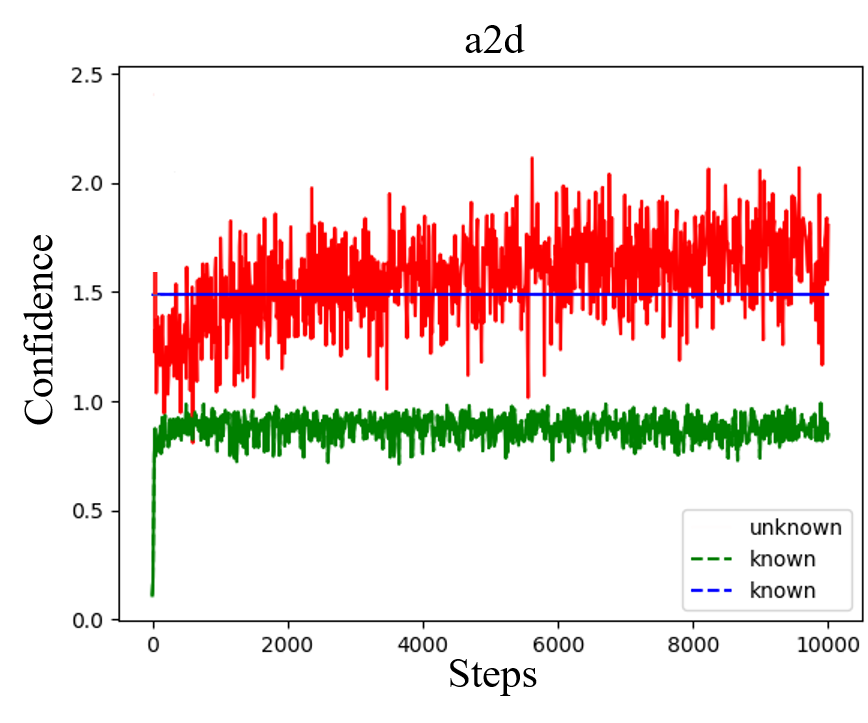}
\end{minipage}%
\begin{minipage}[t]{0.2\linewidth}
\centering
\includegraphics[width=1.45in]{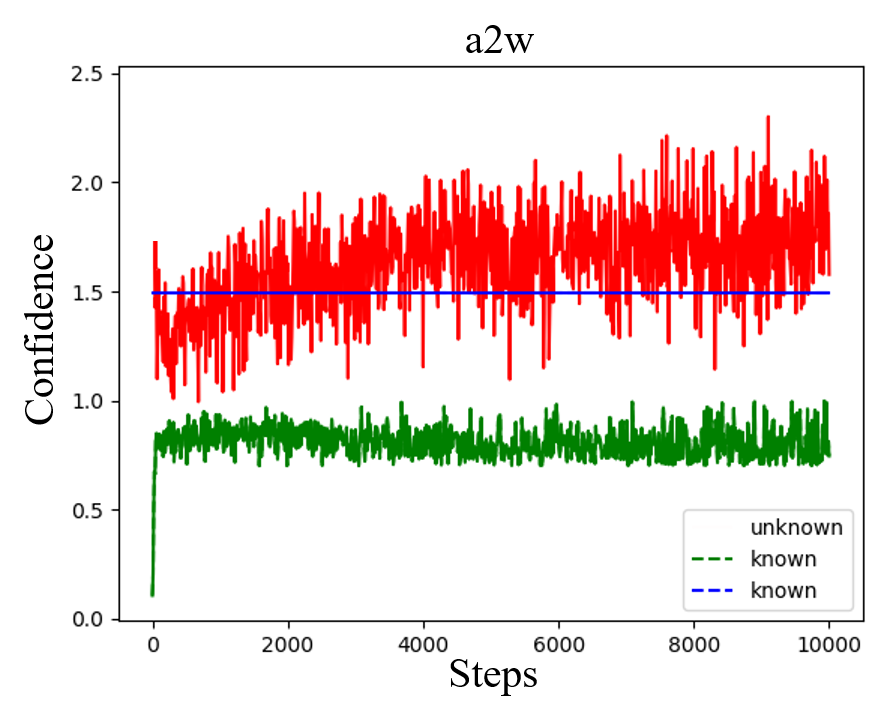}
\end{minipage}%
\begin{minipage}[t]{0.2\linewidth}
\centering
\includegraphics[width=1.45in]{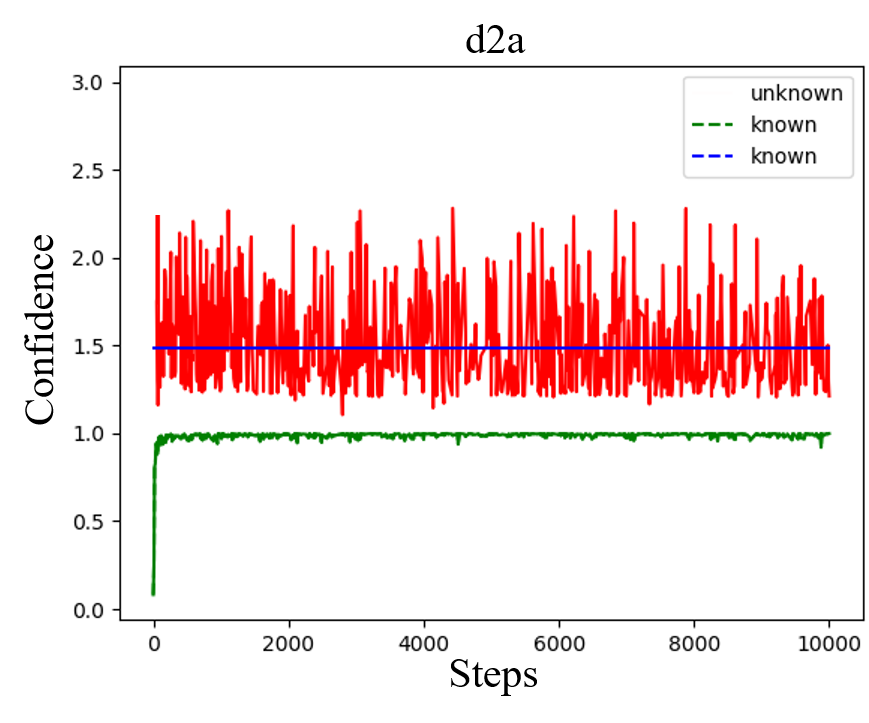}
\end{minipage}%
\begin{minipage}[t]{0.2\linewidth}
\centering
\includegraphics[width=1.45in]{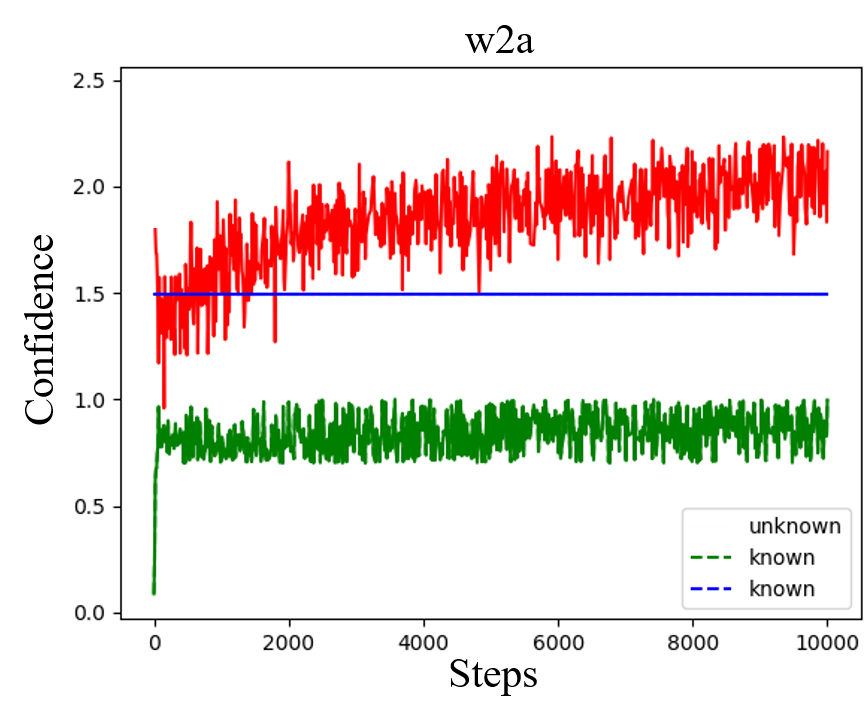}
\end{minipage}%
\begin{minipage}[t]{0.2\linewidth}
\centering
\includegraphics[width=1.45in]{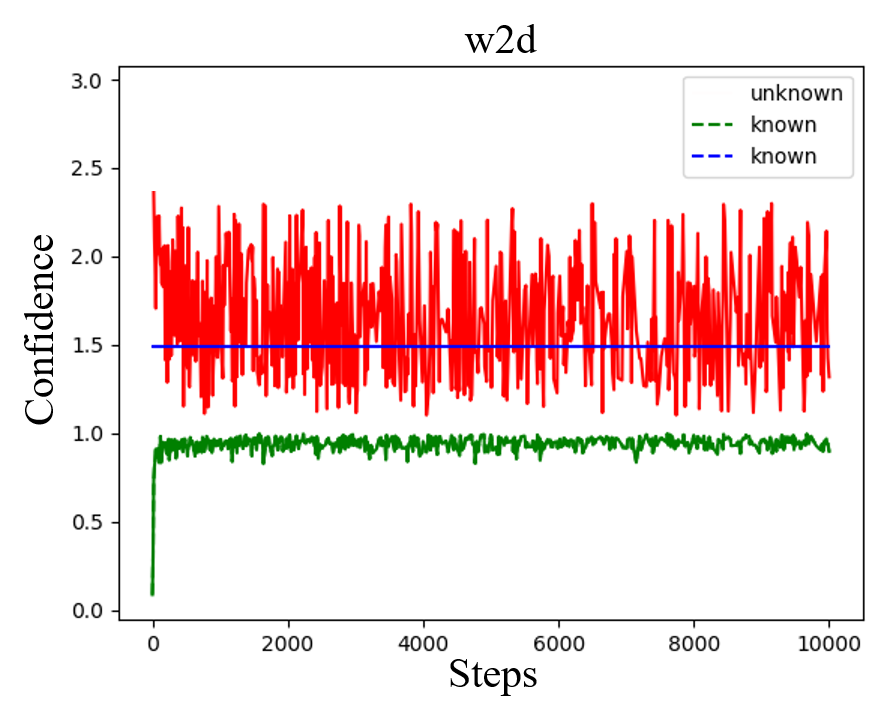}
\end{minipage}%
}%
\\
\subfigure{
\begin{minipage}[t]{0.2\linewidth}
\centering
\includegraphics[width=1.45in]{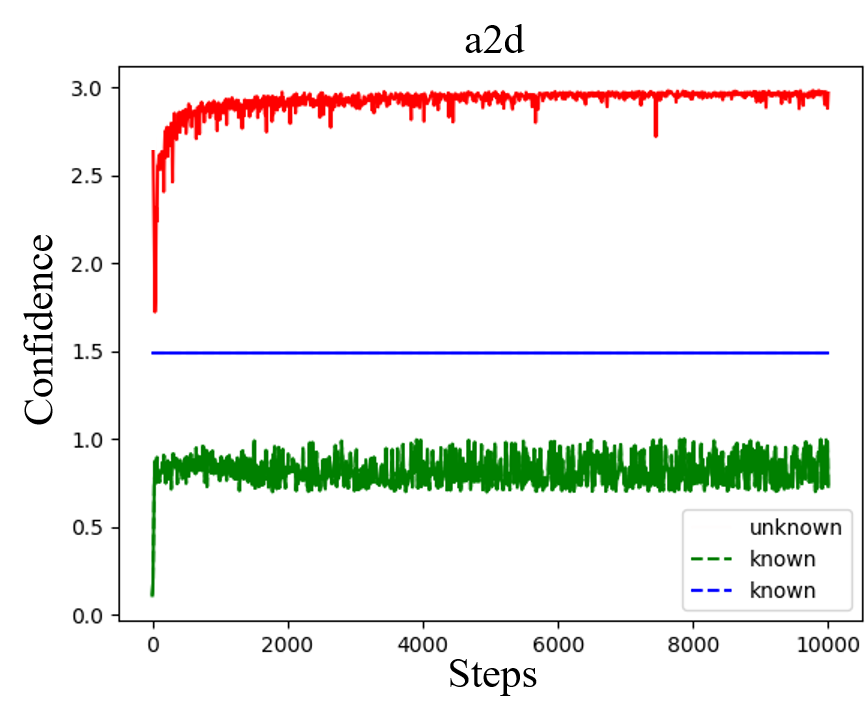}
\end{minipage}%
\begin{minipage}[t]{0.2\linewidth}
\centering
\includegraphics[width=1.45in]{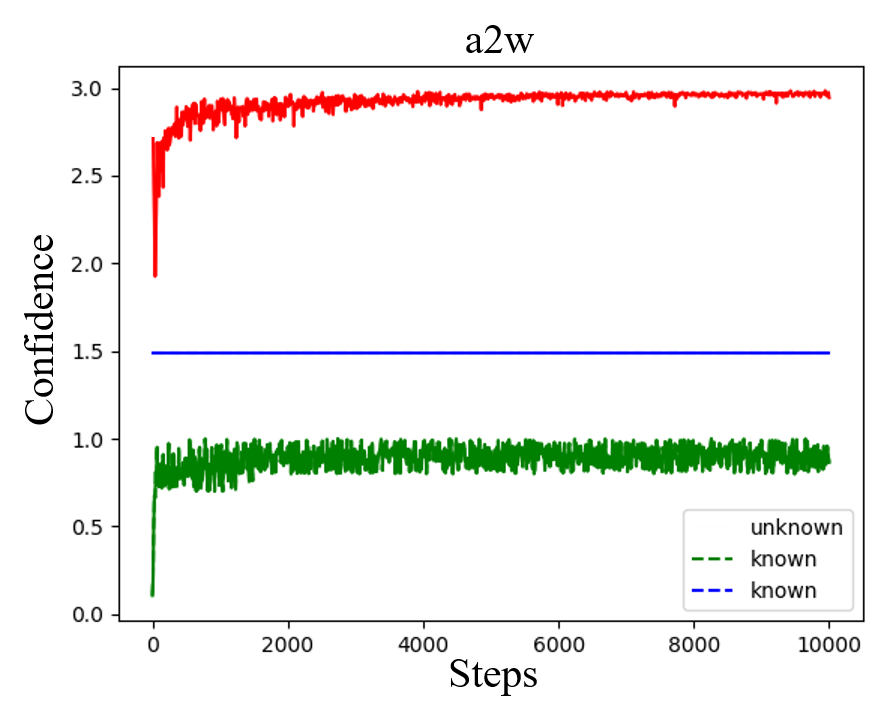}
\end{minipage}%
\begin{minipage}[t]{0.2\linewidth}
\centering
\includegraphics[width=1.45in]{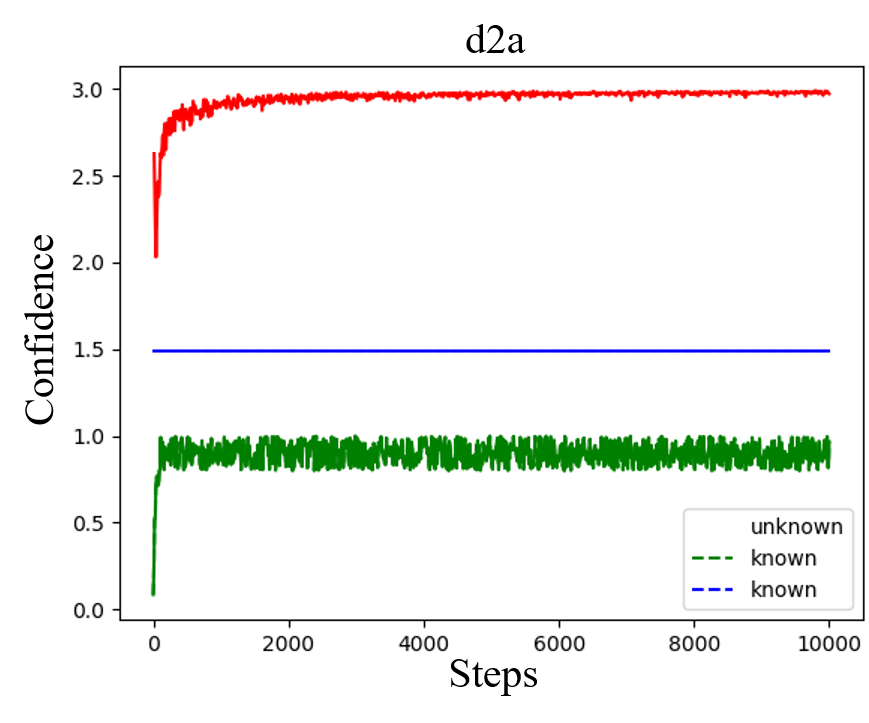}
\end{minipage}%
\begin{minipage}[t]{0.2\linewidth}
\centering
\includegraphics[width=1.45in]{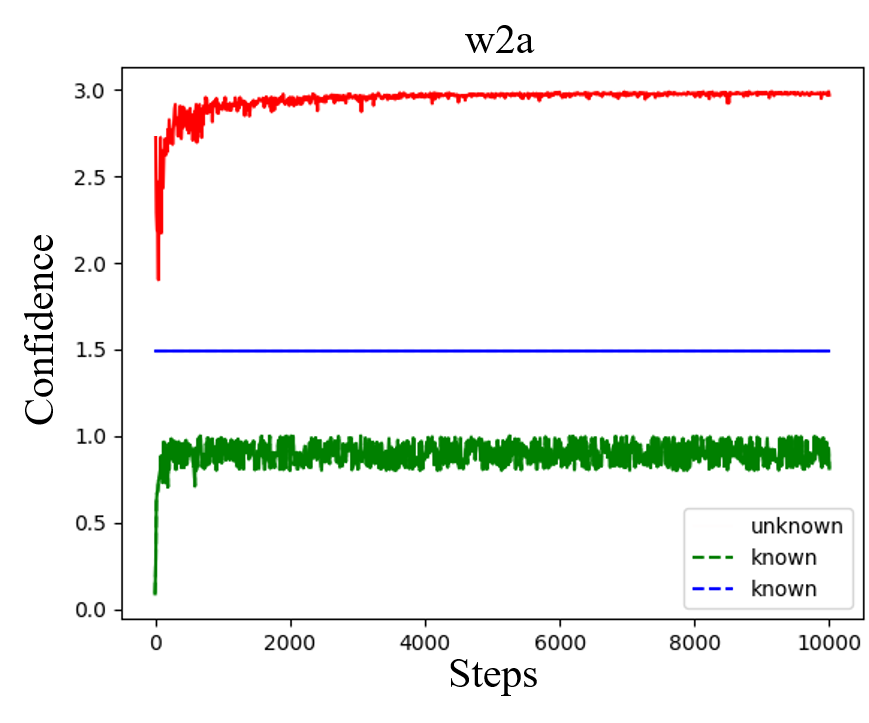}
\end{minipage}%
\begin{minipage}[t]{0.2\linewidth}
\centering
\includegraphics[width=1.45in]{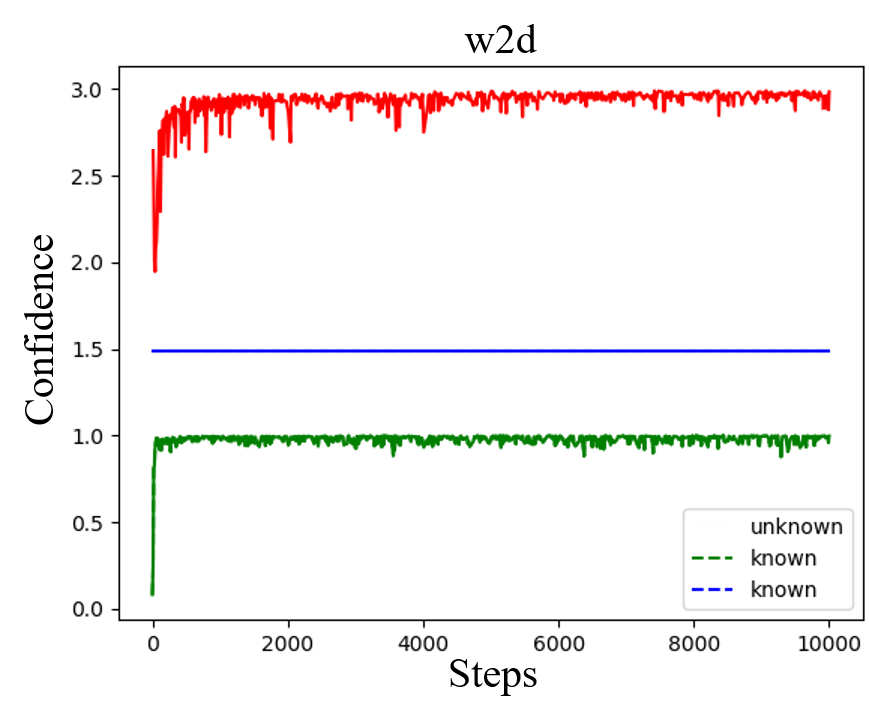}
\end{minipage}%
}%

\vspace{-0.45cm}
\centering
\caption{ Records of entropy of predictions of unknown samples and the maximum values of predictions of known samples following the training process on Office-$31$ with the OPDA setting. The first row represents the distributions of entropy where the model is trained on the source domain only with a CE-loss, the second row represents that where the model is trained with CE-loss and $\mathcal{L}_{unk}$ and the third row represents that with the whole model. The \textcolor{red}{red} lines represent the record of unknown target samples while the \textcolor{green}{green} lines represent that of known target samples.}
\label{ent}
\end{figure*}

\subsubsection{Justification of unknown samples discovering scheme based on the uncertainty estimation}
{\bf Original feature space vs Linear subspace.\ \ } The purpose of extracting a linear subspace is to make the distribution of data points more uniform so as to avoid the influence of the domain misalignment. Comparing the charts in the first row and the second row in Fig.~\ref{kdis}, it is obvious that the data points in the subspace are similar in the distribution of the $k$-nearest neighbor distances. In Fig.~\ref{con}, we conduct experiments on Office-$31$ to compare the distributions of the confidences produced by the proposed  {unknown samples discovering} method in  the original feature space (First row) and  the linear subspace  (Second row). Apparently, the distributions of confidences in the subspace are much more distinguishable than that in the original feature space. The overlaps between unknown and known samples are much fewer in the subspace which is benefited by reducing the misalignment between target samples and source samples.

\begin{figure*}[t]
  \centering
   \includegraphics[width=0.95\linewidth]{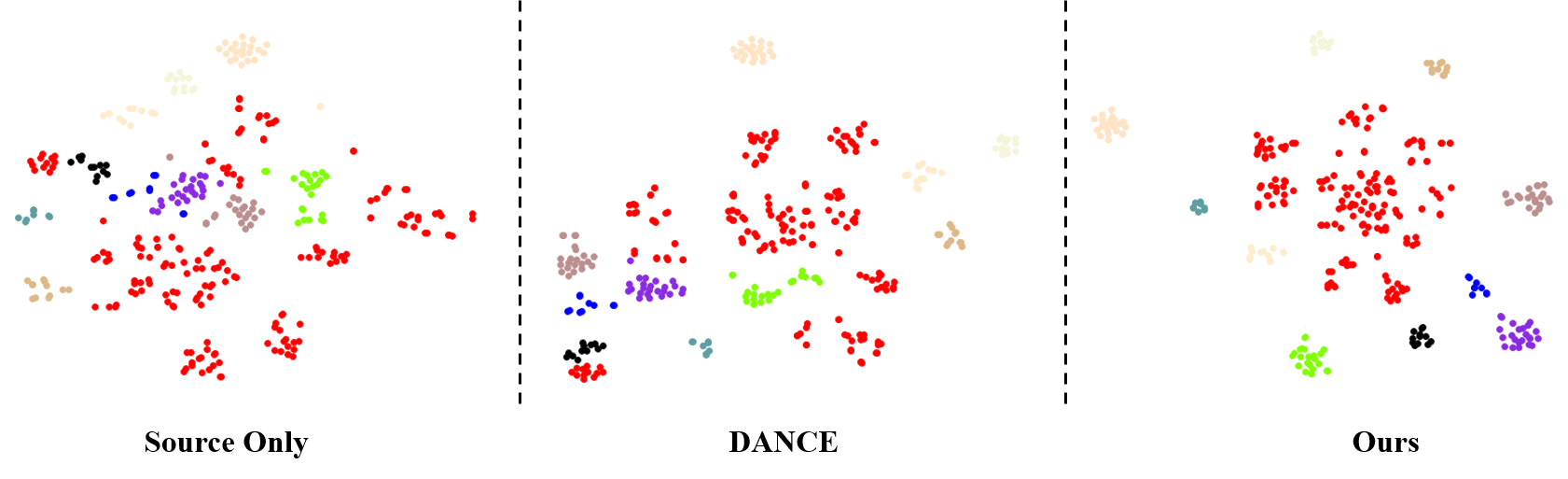}
   \vspace{-1mm}
   \caption{t-SNE visualization on Office-$31$ (A$2$D). Different colors represent different classes. \textcolor{red}{Red} points represent the unknown samples while the points in other colors represent the known samples of different classes. }
   \label{tsne}
\end{figure*}

{\bf Accuracy of the uncertainty estimation.} We also conduct experiments on the accuracy of the uncertainty estimation on the original feature space and the subspace in some subtasks of Office-$31$ and  OfficeHome where we
plot the histograms of the accuracy of the uncertainty estimation in Fig.~\ref{acc}. The unknown samples are consistently detected with high accuracy which on average far surpasses $80\%$  and the performance of accuracy of {unknown samples discovering method based on the uncertainty estimation} in the subspace is much better than that in the original feature space. Thus, through the proposed {unknown samples discovering} scheme, our approach reliably finds the unknown samples in the target domain.

\begin{figure}[t]
\centering
\vspace{-0.3cm}

\subfigure{
\begin{minipage}[t]{0.5\linewidth}
\centering
\includegraphics[width=1.7in]{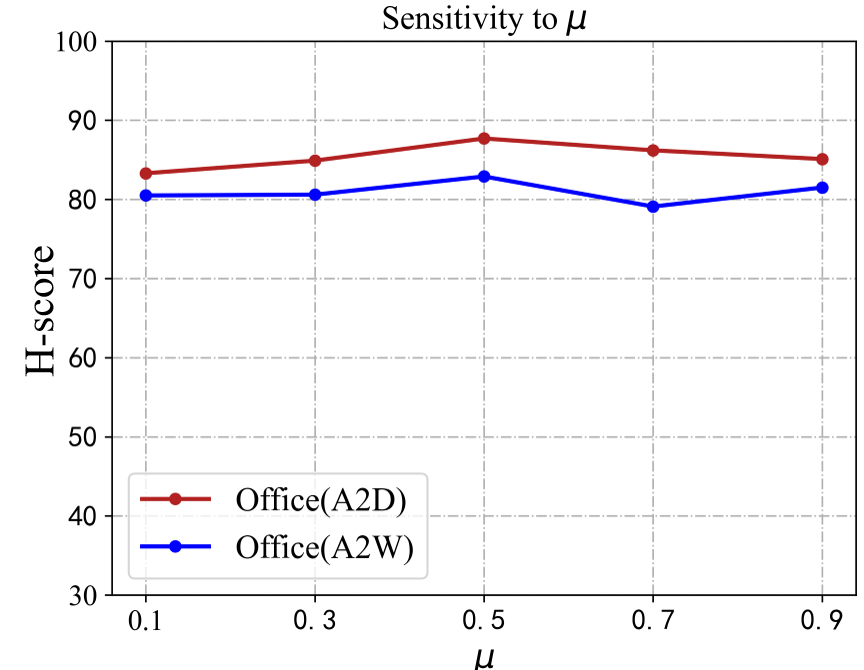}
\end{minipage}%
}%
\subfigure{
\begin{minipage}[t]{0.5\linewidth}
\centering
\includegraphics[width=1.7in]{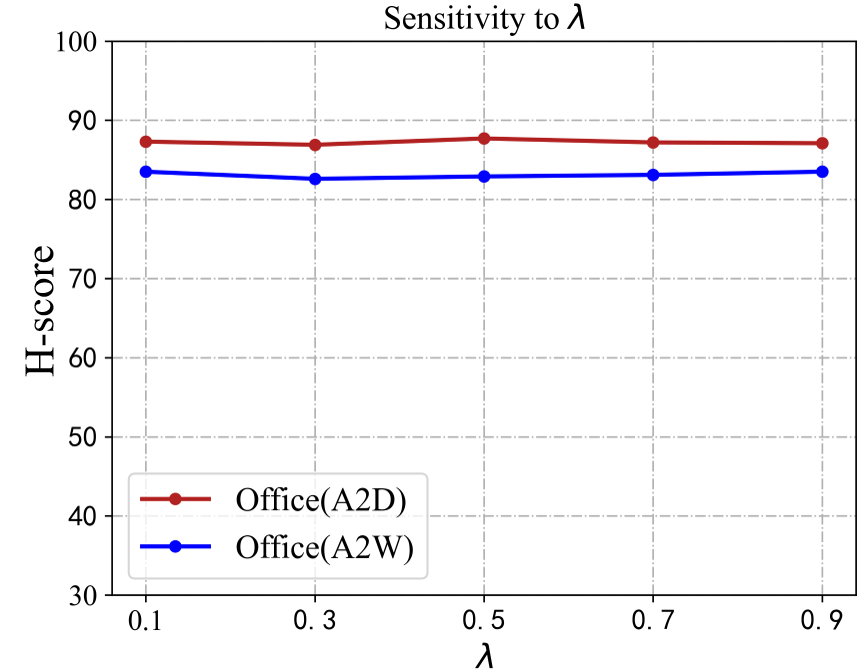}
\end{minipage}%
}%
\vspace{-0.25cm}

\centering
\caption{ {\bf (a)} Results of different human-picked margins in terms of H-score. {\bf (b)} Sensitivity to $\lambda$ in terms of H-score.}
\label{thre_lam}
\end{figure}

{\bf Quantitative comparison with different methods.} To show the improvement on the distribution of the entropy which is used to classify the unknown samples in the test stage, we conducted experiments on Office-31 (a2d). First, we plot the distributions of the entropies of all samples in the target domain at the final epoch in Fig.~\ref{ed}{\bf (a)}. Then, we compare the plot to that trained on the source domain only in Fig.~\ref{ed}{\bf (b)}. We can observe that the full version of our method better distinguishes the known samples from the unknown ones. Furthermore, in Fig.~\ref{ed}{\bf (c)}, we show the corresponding plot produced by DANCE~\cite{saito2020universal} for comparison. Noticeably, our method performs better than DANCE~\cite{saito2020universal} in terms of distinguishing the known samples from the unknown ones.

\subsubsection{Effect of losses}

{\bf Uncertainty-guided margin loss vs CE-loss.\ \ } To show the effect of the uncertainty-guided margin loss on balancing the predictions of known and unknown samples, we track the entropy level of the unknown samples and the confidence level of known samples following the training process on Office-$31$. We recorded the mean value of entropies of predictions for unknown samples and the mean value of maximum prediction confidences output by the classifier of each known sample in every step. For comparison, we first plot records where the classifier was only trained on source domain with a traditional CE-loss like Eq.~(\ref{celoss}) in the first row of Fig.~\ref{ent}. We can observe that although the known samples consistently have high confidence, most of the unknown samples are significantly overconfident during the training process. In the second row of Fig.~\ref{con},  we plot the records using the CE-loss and the proposed unknown loss as Eq.~(\ref{lossunk}). The entropy level has been improved but the overconfidence of the unknown samples is still obvious. In the third row of Fig.~\ref{ent}, we plot the records using the proposed uncertainty-guided loss which can perfectly distinguish the unknown samples using entropy while retaining the known samples with highly confident predictions.
We also employ t-SNE~\cite{van2008visualizing} pictures to visualize the distributions of target samples on Office-$31$ (A$2$D) in Fig.~\ref{tsne}. We observe that the distribution of data points in ours (right) is much more discriminative  than that of DANCE~\cite{saito2020universal} (mid) and the model trained with the source dataset only (left).


{\bf Setting of uncertainty-guided margin. \ \ }
To show the effect of the uncertainty-guided margin selection scheme, we compare it using the human-picked thresholds on Office-$31$ (A$2$D and D$2$A). From Fig.~\ref{thre_lam}{\bf (a)}, we observe that it is difficult to choose a consistently optimal threshold for all datasets and subtasks as the model is sensitive to the thresholds. 

{\bf Different ablated versions of our method.\ \ } Finally, we also provide an ablation study to investigate the effect of each loss in our UniDA framework and show the results in TABLE \ref{tab:abl}. We can see that all losses contribute to the improvement of the results. In particular, among the three target-domain losses, both $\mathcal{L}_{ugm}$ and $\mathcal{L}_{unk}$ have a large impact on the final performance, which demonstrates that it is very important to balance the predictions of known/unknowns samples. 



\subsubsection{Performance on VGGNet}

TABLE~\ref{tab:vgg} shows the quantitative comparison with the ODA setting on Office-31 using VGGNet\cite{simonyan2014very} instead of ResNet-50 as the backbone for feature extraction. According to the results, we demonstrate that our method is also effective with another backbone without changing any hyper-parameters.

\subsubsection{Sensitivity to $\lambda$}

There is only one hyper-parameter $\lambda$ in the loss items. To show the sensitivity of $\lambda$ in the total loss, we conducted experiments on Office-31 (A$2$D and D$2$A) with the OPDA setting. Fig.~\ref{thre_lam}{\bf (b)} shows that our method has a highly stable performance over different values of $\lambda$.
\begin{table}[t]
    \centering
\caption{{Results of different ablated versions of our method on Office-31.}}
    \resizebox{0.95\textwidth}{12mm}{
   \begin{tabular}{c|cccccc|c}
   \hline
    \multirow{2}*{ Method } & \multicolumn{6}{c|}{ Office-31 $(10 / 10 / 11)$} & \multicolumn{1}{l}{} \\
    & A2D & A2W & D2A & D2W & W2D & W2A & Avg \\
    \hline 
     w/o $\mathcal{L}_{ugm}$ & $29. 2$ & $33.4$ & $31.3$ & $52.5$ & $44.2$ & $27.9$ & $36.4$\\
     w/o $\mathcal{L}_{unk}$ & $81.0$ & $77.5$ & $78.2$ & $9 5 . 0$ & $91.0$ & $72.9$ & $82.6$ \\
     w/o $\mathcal{L}_{sup}$ & $8 6 . 9$ & $76.6$ & $8 4.4$ & $91.4$ & $93.3$ & $85.6$ & $8 6 . 3$\\
    \hline
    Ours  & $8 7 . 3$ & $8 3 . 5$ & $82 . 2$ & $96.1$ & $99.2$ & $8 4 . 7$ & $8 8 . 8$\\
    \hline
    \end{tabular}}
    \label{tab:abl}
\end{table}

\begin{table}[t]
    \centering
 \resizebox{0.95\textwidth}{12mm}{
    \begin{tabular}{c|cccccc|c}
   \hline
    \multirow{2}*{ Method } & \multicolumn{6}{c|}{ Office-31 $(10 / 10 / 11)$} & \multicolumn{1}{l}{} \\
    & A2D & A2W & D2A & D2W & W2D & W2A & Avg \\
    \hline OSBP \cite{saito2018open} & $81.0$ & $77.5$ & $78.2$ & $\mathbf{9 5 . 0}$ & $91.0$ & $72.9$ & $82.6$ \\
    ROS \cite{bucci2020effectiveness} & $79.0$ & $81.0$ & $78.1$ & $94.4$ & $\mathbf{9 9 . 7}$ & $74.1$ & $84.4$ \\
    OVANet \cite{saito2021ovanet} & $\mathbf{89.5}$ & $8 4 . 9$ & $8 9 . 7$ & $93.7$ & $85.8$ & $8 8 . 5$ & $8 8 . 7$\\
    \hline
    \rowcolor{gray!30} Ours & $89.4$ & $\mathbf{85 . 6}$ & $\mathbf{92.4}$ & $94.5$ & $90.5$ & $\mathbf{92.2}$ & $\mathbf{90.8}$ \\
   \hline
    \end{tabular}}
    \caption{Results on Office-31 using the VGGNet~\cite{simonyan2014very} backbone with the ODA setting.}
    \label{tab:vgg}
\end{table}
\section{Conclusion}

In this paper, we propose a new framework to reduce the influence of the domain misalignment and balance the predictions of known and unknown target samples. Its core idea is to estimate the probabilities of target samples belonging to the unknown class by the largest number of neighbors with the same label searched from the source domain, and detect the unknown samples via mapping the features in the original feature space into a linear subspace to  reduce the influence of domain misalignment. Also, our method  balances well the confidences of known target samples and unknown target samples via an uncertainty-guided margin loss. As demonstrated by extensive experiments, our method sets the new SOTA performance in various subtasks on three public datasets.


%





\ifCLASSOPTIONcaptionsoff
  \newpage
\fi



\bibliographystyle{IEEEtran}
\bibliography{IEEEabrv,egbib}


%



%





\end{document}